\documentclass[letterpaper]{article} 

\usepackage[preprint]{aaai2027}               

\usepackage[hyphens]{url}  
\usepackage{graphicx} 
\def\UrlFont{\rm}  
\usepackage{natbib}  
\usepackage{caption} 
\usepackage{algorithm}
\usepackage{algorithmic}

\usepackage{newfloat}
\usepackage{listings}

\DeclareCaptionStyle{ruled}{
    labelfont=normalfont,
    labelsep=colon,
    strut=off
} 

\floatstyle{ruled}
\newfloat{listing}{tb}{lst}{}
\floatname{listing}{Listing}

\usepackage{booktabs}
\usepackage{tabularx}
\usepackage{colortbl}

\usepackage{amsmath,amssymb}
\usepackage{multirow}

\graphicspath{{figures/}}

\title{
Continuous-Latent Predictive Modeling with Semantic Alignment
for EEG-Language Foundation Models
}

\author{
Myeong-Ju Cho\textsuperscript{\rm 1},
Hye-Bin Shin\textsuperscript{\rm 1},
Seo-Hyun Lee\textsuperscript{\rm 1},
Seong-Whan Lee\textsuperscript{\rm 2,\textdagger}
}

\affiliations{
\textsuperscript{\rm 1}
Department of Brain and Cognitive Engineering,
Korea University, Seoul, Republic of Korea\\
\textsuperscript{\rm 2}
Department of Artificial Intelligence,
Korea University, Seoul, Republic of Korea\\
\{mj\_cho, hb\_shin, seohyunlee, sw.lee\}@korea.ac.kr
}

\begin{document}

\maketitle


\begingroup
\renewcommand{\thefootnote}{\fnsymbol{footnote}}
\footnotetext[2]{Corresponding author}
\endgroup


\begin{abstract}
Recent advances in EEG foundation models have demonstrated the potential
of large-scale pretraining to enable generalizable neural decoding across
subjects, recording environments, and datasets. However, dominant
pretraining paradigms face key challenges: masked autoencoding tends to
prioritize low-level signal reconstruction over task-relevant semantics,
while autoregressive modeling creates a mismatch between continuous neural
dynamics and discrete token spaces. To address these challenges, new
strategies are needed to effectively align continuous EEG representations
with natural-language semantics and enable their integration with large
language models. Accordingly, we propose \textbf{Brain Latent Predictive
Model (BLPM)}, an EEG--language foundation model that reformulates
heterogeneous EEG decoding tasks as a continuous semantic embedding
prediction problem. BLPM introduces a Continuous EEG Latent Predictive
(CELP) encoder that learns transferable representations through latent
target prediction. Building on these representations, a Multi-Query
Semantic Decomposition (MQSD) module extracts task-relevant information
and aligns continuous EEG representations with textual semantics within a
shared latent space according to their semantic relationships. Experiments
across multiple benchmarks demonstrate consistent generalization
performance across diverse tasks, establishing continuous latent semantic
prediction as an effective paradigm for EEG-language foundation models.
\end{abstract}


\section{Introduction}

Foundation models have achieved remarkable advances across diverse modalities, including vision \cite{oquab2023dinov2}, language \cite{radford2019language}, and speech \cite{radford2023robust}. Their integration with pretrained large language models (LLMs) has further advanced multimodal learning and expanded their applications to healthcare domains such as clinical diagnosis \cite{dai2026qoq} and biosignal analysis \cite{gu2026cerebragloss}. Motivated by these innovative advances, electroencephalography (EEG) decoding research has increasingly shifted from task-specific models toward universal foundation models designed to generalize across different subjects, recording environments, and datasets. This transition has been driven by self-supervised learning (SSL) on large-scale unlabeled EEG data, enabling EEG foundation models to learn transferable representations that can be applied to diverse downstream tasks within a single unified model \cite{jiang2024labram,wang2025cbramod}. Despite this progress, existing approaches still retain several fundamental limitations.

Prior studies on EEG foundation models mainly learn universal representations through masked signal reconstruction \cite{wang2025cbramod, zhou2026csbrain, elouahidi2025reve}, masked discrete token prediction \cite{jiang2024labram, pradeepkumar2026tokenizing, ma2026codebrain}, or autoregressive next-token prediction \cite{jiang2025neurolm, yang2025thdbar, wang2026kastbar}. Their training and evaluation rely predominantly on reconstruction and prediction objectives defined either in the input space or in discrete representation spaces.

Among these paradigms, masked autoencoding \cite{he2022masked} tends to focus on reconstructing low-level signal structures. The learned representations can remain sensitive to fine-grained waveform characteristics, noise, and subject-specific variability \cite {wu2024neuro}. Moreover, they do not explicitly organize EEG representations around task-relevant semantic concepts or directly align them with the semantic space of pretrained LLMs. Consequently, strong reconstruction performance does not necessarily guarantee semantically rich and language-compatible EEG representations that generalize across heterogeneous EEG decoding tasks. 

Meanwhile, LLMs are inherently designed to model discrete token sequences, such as text. Therefore, converting continuous EEG embeddings into discrete neural tokens and modeling them through autoregressive next-token prediction has been established as a prominent paradigm for integrating EEG signals with LLMs \cite{jiang2025neurolm,yang2025thdbar,wang2026kastbar}. While this formulation enables the use of established sequence-modeling frameworks, quantizing continuous neural dynamics into a finite codebook may discard discriminative patterns that are critical for distinguishing task-relevant neural states \cite{cui2026brainrvq, zhai2026vp}. Moreover, converting all neural information into discrete tokens and optimizing for next-token prediction can also create an unnecessary representational bottleneck \cite{zheng2025rethinking}. Autoregressive modeling is well suited for text generation, but it may not be the most appropriate learning objective when the primary goal is discriminative EEG decoding guided by semantic understanding. These limitations motivate alternative approaches that directly learn semantic structure in continuous latent spaces without being constrained by discrete tokenization or an autoregressive next-token prediction objective.

To address these challenges, we propose the \textbf{Brain Latent Predictive Model (BLPM)}, a non-generative and non-autoregressive EEG–language foundation model that reformulates heterogeneous EEG decoding tasks as a continuous semantic embedding prediction problem in a shared latent space. Rather than directly reconstructing raw EEG waveforms, BLPM employs a \textbf{Continuous EEG Latent Predictive (CELP) Encoder} that predicts latent representations of target EEG segments from contextual EEG observations, thereby reducing excessive dependence on low-level waveform details and subject-specific variations while learning transferable representations that capture higher-level neurophysiological structures. BLPM then connects the learned continuous EEG representations to a text semantic embedding space without converting EEG signals into discrete neural tokens. To this end, we introduce a \textbf{Multi-Query Semantic Decomposition (MQSD) module}, which uses multiple language-guided semantic queries to extract complementary task-relevant information from each EEG representation. This semantic information is then integrated with the continuous EEG embeddings to predict the embedding of the corresponding answer in a shared latent space. The predicted embedding is matched against the embeddings of candidate answers, allowing the model to select the answer that is semantically most compatible with the EEG representation. This design enables distinct semantic factors contained within the same EEG segment to be selectively accessed according to the given task, thereby reformulating heterogeneous EEG decoding tasks within a unified latent semantic prediction framework.

The main contributions of this paper are as follows:

\begin{itemize}

\item
We propose BLPM, an EEG-language foundation model that aligns continuous EEG representations with language semantics through semantic embedding prediction in a latent space. This formulation enables unified EEG decoding without discrete neural tokenization or autoregressive generation.

\item
We introduce the CELP encoder, which learns universal EEG representations by predicting target latent representations from contextual EEG observations. Its latent predictive objective promotes higher-level abstraction without directly reconstructing low-level details.

\item
We present the MQSD module, which decomposes continuous EEG representations into multiple task-relevant semantic queries with language-derived semantic guidance. Combined with multi-task instruction tuning and semantic answer matching, it reformulates heterogeneous EEG decoding tasks as unified latent semantic prediction problems.

\item
We systematically evaluate BLPM using standardized benchmarks to enable fair comparisons and demonstrate its robust performance. These results support continuous latent semantic prediction as an effective paradigm for EEG-language foundation models.

\end{itemize}

\begin{figure*}[!t]
\centering
\includegraphics[width=\textwidth]{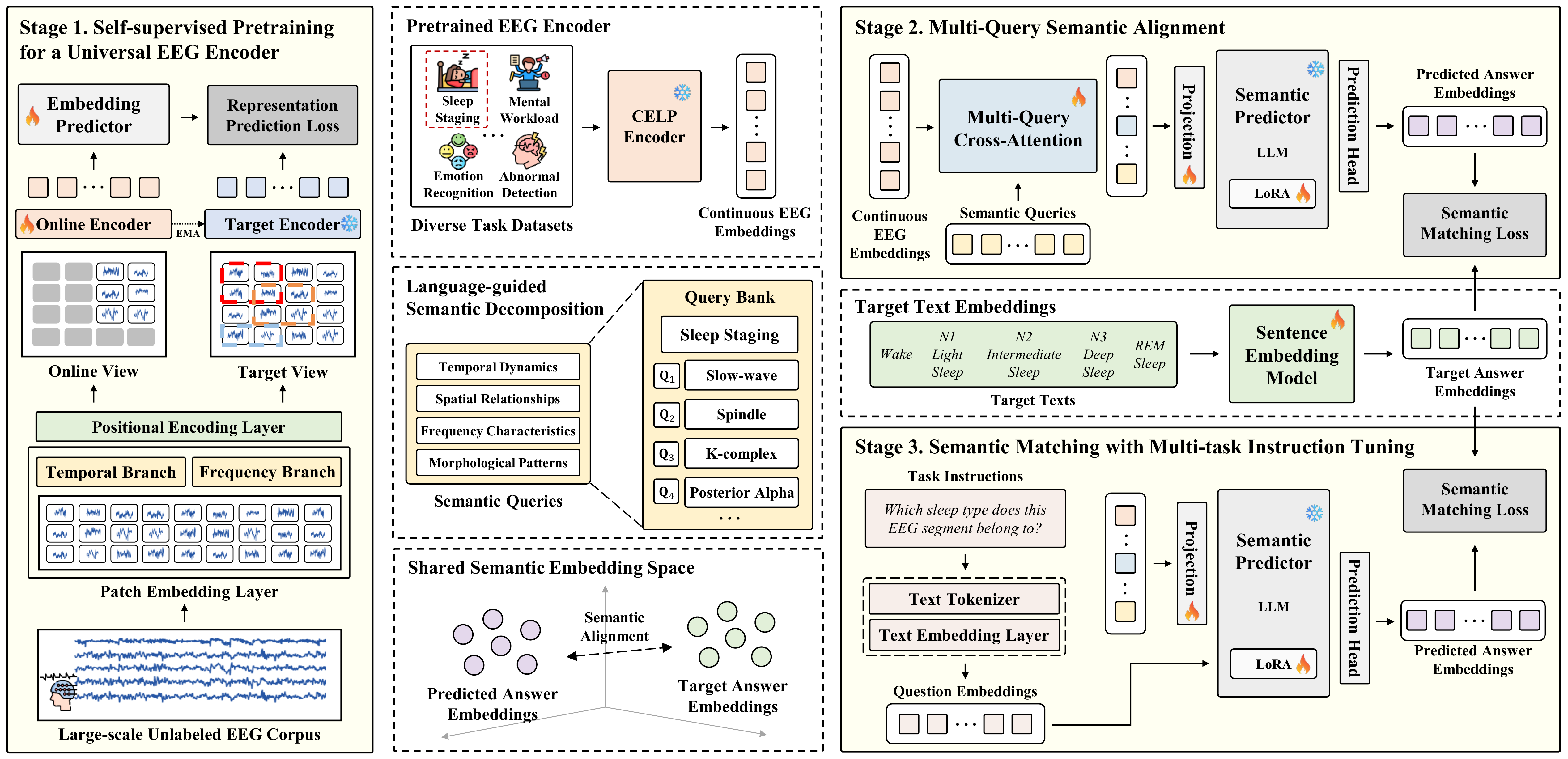}
\caption{Overall framework of the BLPM. Stage 1 trains the CELP encoder through continuous latent representation prediction. Stage 2 aligns continuous EEG representations with natural-language semantics through language-guided multi-query decomposition, while Stage 3 performs instruction-conditioned answer embedding prediction and semantic matching.}
\label{fig:overall_framework2}
\end{figure*}

\section{Methods}
In this section, we introduce BLPM, a continuous latent predictive EEG–language foundation model that aligns EEG representations with natural language semantics in a shared latent space. As illustrated in Figure 1, BLPM comprises three main components: (1) a CELP encoder for learning universal continuous EEG representations; (2) a Language-Guided MQSD module for aligning EEG representations with language semantics; and (3) a multi-task instruction-tuning framework with semantic answer matching for unified EEG decoding.

\subsection{Continuous EEG Latent Predictive Encoder}

\subsubsection{Input Formulation.}
Given input EEG data $X_{\mathrm{EEG}} \in \mathbb{R}^{C \times T}$, we divide each channel into non-overlapping segments of length $L$, obtaining $X \in \mathbb{R}^{C \times P \times L}$, where $C$ is the number of channels, $T$ is the number of temporal samples, and $P = \lfloor T/L \rfloor$ is the number of segments per channel. Each segment $x_{c,p} \in \mathbb{R}^{L}$ where $c \in \{1, \dots, C\}$ and $p \in \{1, \dots, P\}$ is treated as an individual EEG token, resulting in a total sequence of $N = C \times P$ tokens.

\subsubsection{Patch Embedding.}
Each EEG token $x_{c,p} \in \mathbb{R}^{L}$ is mapped into a $d$-dimensional latent vector through a dual-branch patch embedding module. The temporal branch applies 1D convolutional layers to the segment to capture local temporal dynamics, while the spectral branch computes the fast Fourier transform (FFT) of $x_{c,p}$ and projects its magnitude spectrum into the same $d$-dimensional space via a linear layer. The temporal and spectral embeddings are additively fused to form spatio-temporal patch embeddings $Z \in \mathbb{R}^{C \times P \times d}$. A depthwise convolutional positional encoding is then applied across the channel and temporal dimensions to inject spatio-temporal positional information. Finally, the channel and temporal dimensions are flattened into a single token dimension, producing the position-encoded token embeddings $\tilde{Z} \in \mathbb{R}^{N \times d}$.

\subsubsection{Joint Latent Predictive Pretraining.}
The CELP Encoder is trained with a latent predictive pretraining scheme inspired by the joint-embedding predictive architecture (JEPA) \cite{assran2023self}. The scheme consists of an online encoder, a target encoder, and an embedding predictor.

\subsubsection{Structured Multi-Block Masking.}
To preserve the intrinsic structure of EEG signals, the CELP encoder applies multi-block masking over the channel-time EEG token sequence. The visible mask $M_v$ defines the observable context tokens, while the prediction masks $\{M_p\}$ define structured target regions to be inferred. In practice, the target masks include temporal-span masks and channel-block masks, encouraging the model to exploit temporal continuity, channel-wise dependencies, and inter-token context.

\subsubsection{Online Encoder.}
The online encoder $f_{\theta}$ takes the visible subset of token embeddings $\tilde{\mathbf{Z}}_{M_v}$ as input and encodes them into visible latent representations $\mathbf{H}_v$:
\begin{equation}
\mathbf{H}_v = f_{\theta}(\tilde{\mathbf{Z}}_{M_v}).
\label{eq:online_encoder}
\end{equation}
Unlike masked autoencoding, which reconstructs masked segments, the CELP encoder predicts masked targets in the latent space from visible tokens. This design encourages the model to learn abstract semantic representations rather than relying on reconstruction objectives that are sensitive to low-level noise and fluctuations.

\subsubsection{Target Encoder.}
The target encoder $f_{\bar{\theta}}$ processes the full EEG token sequence and provides latent targets at masked prediction positions. At training step $k$, the target encoder parameters $\bar{\theta}$ are updated using the exponential moving average (EMA) of the online encoder parameters $\theta$ with a momentum coefficient $m_k$:
\begin{equation}
\bar{\theta}^{(k)}
\leftarrow
m_k \bar{\theta}^{(k-1)}
+
(1-m_k)\theta^{(k)}.
\label{eq:ema_update}
\end{equation}
This EMA encoder provides stable latent prediction targets and helps mitigate representation collapse during pretraining.

\subsubsection{Embedding Predictor.}
For each prediction mask $M_p$, the embedding predictor $g_{\phi}$ estimates the masked latent representations from the visible context and learnable mask queries:
\begin{equation}
\hat{\mathbf{H}}_{M_p}
=
g_{\phi}
\left(
\mathbf{H}_v,
\mathbf{m}+\mathbf{p}_{M_p}
\right).
\label{eq:embedding_predictor}
\end{equation}
where $\mathbf{m}$ is a learnable mask token and $\mathbf{p}_{M_p}$ denotes the positional representation of the masked target positions. Once pretraining is completed, the online encoder is used as the pretrained EEG encoder backbone.

\subsubsection{Latent Predictive Objective.}
The training objective is defined in the latent space rather than the input space. Let $\mathbf{H}_t = f_{\bar{\theta}}(\tilde{\mathbf{Z}})$ denote the target latent representations obtained from the target encoder. The target representations are normalized along the feature dimension and used as prediction targets. The CELP encoder is optimized with a Smooth L1 loss over masked positions:
\begin{equation}
\mathcal{L}_{\mathrm{CELP}}
=
\frac{1}{|\mathcal{M}|}
\sum_{M_p \in \mathcal{M}}
\frac{1}{|M_p|}
\sum_{i \in M_p}
\rho
\left(
\hat{\mathbf{h}}_i,
\mathrm{LN}(\mathbf{h}_{t,i})
\right).
\label{eq:celp_objective}
\end{equation}
where $\rho(\cdot,\cdot)$ denotes the Smooth L1 loss, $\mathcal{M}$ is the set of prediction masks, $\hat{\mathbf{h}}_i$ and $\mathbf{h}_{t,i}$ denote the predicted and target latent vectors at position $i$, and $\mathrm{LN}(\cdot)$ represents layer normalization.

After pretraining, the CELP encoder is applied to the token embeddings $\tilde{\mathbf{Z}}$, and the output sequence is reshaped into continuous EEG embeddings:
\begin{equation}
\mathbf{H}_{\mathrm{EEG}}
=
\operatorname{reshape}
\left(
f_{\theta}(\tilde{\mathbf{Z}})
\right)
\in
\mathbb{R}^{C \times P \times d}.
\label{eq:continuous_eeg_embeddings}
\end{equation}

\subsection{Multi-Query Semantic Alignment}
In this stage, we freeze the CELP encoder and align its continuous EEG embeddings with natural-language semantics using the \textbf{Language-Guided Multi-Query Semantic Decomposition (MQSD) module} and \textbf{Language-Initialized Semantic Embedding Predictor}. The goal of this stage is to predict natural-language semantic representations in a semantic embedding space from continuous EEG representations.
\subsubsection{Continuous EEG Representations.}
The frozen CELP encoder provides continuous EEG embeddings $\mathbf{H}_{\mathrm{EEG}} \in \mathbb{R}^{C \times P \times d}$. We flatten the channel-time dimensions into a token sequence:
\begin{equation}
\mathbf{h}
=
\mathrm{Flatten}
\left(
\mathbf{H}_{\mathrm{EEG}}
\right)
\in
\mathbb{R}^{N \times d}.
\label{eq:eeg_token_sequence}
\end{equation}
The EEG token sequence $\mathbf{h}$ serves directly as the key and value for the multi-query cross-attention, retaining its original feature space prior to projection.

\subsubsection{Language-Guided Semantic Queries.}
To interpret the EEG representations from complementary perspectives, MQSD introduces a fixed set of $M$ semantic queries. These queries correspond to task-relevant EEG factors, including temporal dynamics, spectral characteristics, spatial relationships, and morphological patterns. Each semantic query is tokenized and embedded using the input embedding layer of the LLM. The $m$-th query is represented by a padded sequence of language token embeddings $\mathbf{Q}_m \in \mathbb{R}^{L_{\max} \times d_{\mathrm{LM}}}$, where $L_{\max}$ is the maximum query length. A corresponding attention mask is used to exclude padding tokens.

\subsubsection{Multi-Query EEG Decomposition.}
Each semantic query sequence independently attends to the continuous EEG tokens through semantic query cross-attention:
\begin{equation}
\begin{aligned}
\mathbf{R}_m
&=
\mathrm{CrossAttn}(\mathbf{Q}_m,\mathbf{h},\mathbf{h}),
\\
\tilde{\mathbf{s}}_m
&=
\mathrm{Pool}_{\mathrm{tok}}(\mathbf{R}_m).
\end{aligned}
\label{eq:multi_query_decomposition}
\end{equation}
Here, $\mathbf{R}_m \in \mathbb{R}^{L_{\max} \times d}$ indicates the query-conditioned EEG representations associated with the language tokens of the $m$-th semantic query. The operator $\mathrm{Pool}_{\mathrm{tok}}$ performs masked mean pooling over the language-token dimension, yielding an intermediate semantic summary $\tilde{\mathbf{s}}_m \in \mathbb{R}^{d}$.

Finally, each semantic summary is projected into the hidden dimension of the LLM using the linear projector $P_h$:
\begin{equation}
\mathbf{s}_m
=
P_h(\tilde{\mathbf{s}}_m)
\in
\mathbb{R}^{d_{\mathrm{LM}}}.
\label{eq:semantic_summary_projection}
\end{equation}
The semantic summaries obtained from all $M$ queries are collected as
\begin{equation}
\mathbf{S}_q
=
[\mathbf{s}_1;\mathbf{s}_2;\ldots;\mathbf{s}_M]
\in
\mathbb{R}^{M \times d_{\mathrm{LM}}}.
\label{eq:semantic_summary_collection}
\end{equation}
This decomposition allows the model to extract complementary semantic summaries from the same continuous EEG embeddings rather than compressing all the information into a single global representation.

\subsubsection{Language-Initialized Semantic Embedding Predictor.}
To preserve fine-grained information alongside the semantic summaries, the continuous EEG embeddings are also projected into the LLM embedding space via a token projector $P_u$:
\begin{equation}
\mathbf{u}
=
P_u(\mathbf{h})
\in
\mathbb{R}^{N \times d_{\mathrm{LM}}}.
\label{eq:eeg_token_projection}
\end{equation}
The multi-query semantic summaries $\mathbf{S}_q$ and projected EEG embeddings $\mathbf{u}$ are then concatenated and fed into the semantic embedding predictor:
\begin{equation}
\mathbf{H}
=
G([\mathbf{S}_q;\mathbf{u}]).
\label{eq:semantic_predictor}
\end{equation}
where $G$ indicates the semantic embedding predictor. The projected EEG embeddings preserve fine-grained token-level EEG information, while $\mathbf{S}_q$ provides language-guided semantic summaries. Although $G$ consists of pretrained LLM decoder layers, we use a bidirectional attention mask instead of a causal mask to predict semantic embeddings rather than to generate the next token.

We extract the contextualized hidden states corresponding to the $M$ semantic-query summaries:
\begin{equation}
\mathbf{H}_q
=
\mathbf{H}_{1:M}.
\label{eq:semantic_query_states}
\end{equation}
We then obtain the predictive representation by mean pooling over the semantic-query states:
\begin{equation}
\mathbf{h}_{\mathrm{pred}}
=
\mathrm{Pool}_{\mathrm{qry}}(\mathbf{H}_q)
=
\frac{1}{M}
\sum_{m=1}^{M}
\mathbf{H}_{q,m}.
\label{eq:query_pooling}
\end{equation}
The pooled representation is mapped into the text embedding space:
\begin{equation}
\mathbf{z}_e
=
\operatorname{norm}
\left(
P_e(\mathbf{h}_{\mathrm{pred}})
\right).
\label{eq:eeg_semantic_embedding}
\end{equation}
where $P_e$ is the prediction head, and $\operatorname{norm}(\cdot)$ denotes L2 normalization.

\subsubsection{Natural-Language Semantic Targets.}
For the text branch, each class label $y$ is expressed as natural-language text $\ell_y$ representing its class semantics. The semantic text is encoded using a pretrained text embedding model:
\begin{equation}
\mathbf{z}_y
=
\operatorname{norm}
\left(
E_{\mathrm{text}}(\ell_y)
\right).
\label{eq:text_semantic_embedding}
\end{equation}
This maps labels from heterogeneous EEG tasks into a shared natural-language semantic embedding space.

\subsubsection{Semantic Alignment Objective.}
We optimize the model using a bidirectional multi-positive contrastive objective. The similarity between the $i$-th predicted EEG-conditioned embedding $\mathbf{z}_{e,i}$ and the $j$-th target semantic embedding $\mathbf{z}_{y,j}$ is computed as
\begin{equation}
A_{ij}
=
\mathbf{z}_{e,i}^{\top}\mathbf{z}_{y,j}/\tau.
\label{eq:semantic_similarity}
\end{equation}
where $\tau$ is a temperature parameter, and both $\mathbf{z}_e$ and $\mathbf{z}_y$ are L2-normalized embeddings in the shared semantic space. The final objective is a symmetric InfoNCE loss:
\begin{equation}
\mathcal{L}_{\mathrm{align}}
=
\frac{1}{2}
\left[
\mathcal{L}_{e \rightarrow t}
+
\mathcal{L}_{t \rightarrow e}
\right].
\label{eq:alignment_objective}
\end{equation}
Samples sharing the same semantic text are treated as positives, preventing samples with the same class semantics from being used as false negatives.

\subsection{Multi-task Instruction Tuning with Semantic Embedding Prediction}
We perform multi-task instruction tuning to adapt BLPM to diverse downstream datasets under a unified semantic prediction framework. Instead of autoregressively generating answer tokens, BLPM predicts the semantic embedding of the target answer in a single forward pass.

\subsubsection{Instruction-Conditioned Semantic Prediction.}
Given the multi-query semantic summaries $\mathbf{S}_q$ and projected EEG tokens $\mathbf{u}$ obtained in the previous stage, the task instruction $s_k$ is tokenized and embedded using the input embedding layer of the pretrained language model:
\begin{equation}
\mathbf{T}_k
=
\mathrm{Embed}_{\mathrm{LM}}(s_k).
\label{eq:instruction_embedding}
\end{equation}
The instruction embeddings are concatenated with the EEG tokens and semantic summaries, and then processed by the non-causal LLM as a semantic predictor:
\begin{equation}
\mathbf{H}^{(k)}
=
G\left([\mathbf{T}_k;\mathbf{S}_q;\mathbf{u}]\right).
\label{eq:instruction_predictor}
\end{equation}
The predicted answer embedding is obtained from the contextualized semantic-query states using the same query-level pooling employed in the semantic alignment stage:
\begin{equation}
\hat{\mathbf{z}}
=
\operatorname{norm}
\left(
P_a
\left(
\mathrm{Pool}_{\mathrm{qry}}(\mathbf{H}^{(k)}_q)
\right)
\right).
\label{eq:answer_embedding}
\end{equation}
where $\mathbf{H}^{(k)}_q$ denotes the contextualized hidden states corresponding to the $M$ semantic-query summaries, and $P_a$ is the shared answer prediction head. This formulation conditions semantic prediction on the task instruction while preserving a shared prediction architecture across tasks.

\subsubsection{Instruction Tuning Objective.}
Each target label $y$ is associated with the natural-language semantic text $\ell_y$ and its corresponding target embedding $\mathbf{z}_y$. The model is optimized using the bidirectional multi-positive InfoNCE objective \cite{lee2022uniclip} to align the predicted answer embedding $\hat{\mathbf{z}}$ with $\mathbf{z}_y$. Samples with the same answer semantics are treated as positive pairs.

\subsubsection{Inference by Semantic Matching.}
During inference, the candidate answer texts for task $k$ are encoded using the pretrained text embedding model \cite{vera2025embeddinggemma} and projected into the shared semantic embedding space. The predicted label is selected by semantic matching within the task-specific candidate set:
\begin{equation}
\hat{y}
=
\arg\max_{c \in \mathcal{C}_k}
\hat{\mathbf{z}}^{\top}\mathbf{z}_c.
\label{eq:semantic_matching}
\end{equation}
where $\mathcal{C}_k$ denotes the candidate label set for task $k$, and $\mathbf{z}_c$ is the embedding of the natural-language semantic text $\ell_c$ representing candidate label $c$. This formulation unifies heterogeneous EEG tasks within a shared semantic space without relying on discrete tokenization and autoregressive next-token prediction.

\section{Experimental Setup}
\subsection{Datasets and Benchmarks}
\subsubsection{Pretraining.}
BLPM is pretrained on the Temple University Hospital EEG Corpus (TUEG) \cite{obeid2016temple}, a large-scale publicly available clinical EEG dataset comprising 69,652 recordings from 14,987 subjects, with a total of 27,062 hours of recordings.

\subsubsection{Downstream Tasks.}
To assess the generalization ability of BLPM across diverse EEG applications, we conduct experiments on seven representative downstream tasks. The tasks include mental workload classification (COG-BCI~\cite{hinss2023open}), mental stress detection (Mental Arithmetic~\cite{zyma2019workload}), abnormal detection (TUAB~\cite{lopez2015automated}), event type classification (TUEV~\cite{harati2015improved}), motor imagery classification (PhysioNet-MI~\cite{PhysioNet-eegmmidb-1.0.0}), emotion recognition (FACED~\cite{chen2023large}), and sleep staging (HMC~\cite{PhysioNet-hmc-sleep-staging-1.1}).

\subsubsection{Evaluation Benchmark.}
We use NeuralBench \cite{banville2026neuralbench}, a unified framework for benchmarking models of brain activity to evaluate model performance systematically and fairly across various downstream tasks and datasets. NeuralBench provides standardized evaluation protocols, enabling consistent comparisons across datasets, tasks, and model architectures. Unless otherwise specified, we follow the standard benchmark settings for task formulations, data splits, preprocessing, and evaluation metrics.

\begin{table*}[!t]
\centering

{\small
\renewcommand{\arraystretch}{1.06}
\setlength{\tabcolsep}{2pt}

\begin{tabular*}{\textwidth}{
@{\extracolsep{\fill}}
c c l c c c c c c c
@{}
}
\toprule

\multirow{2}{*}{\textbf{Task}}
& \multirow{2}{*}{\textbf{Paradigm}}
& \multirow{2}{*}{\textbf{Method}}
& \multirow{2}{*}{\textbf{COG-BCI}}
& \textbf{Mental}
& \multirow{2}{*}{\textbf{TUAB}}
& \multirow{2}{*}{\textbf{TUEV}}
& \multirow{2}{*}{\textbf{PhysioNet-MI}}
& \multirow{2}{*}{\textbf{FACED}}
& \multirow{2}{*}{\textbf{HMC}}
\\

&
&
&
&
\textbf{Arithmetic}
&
&
&
&
&
\\

\midrule

\multirow{7}{*}{\textit{Single-Task}}
& \multirow{2}{*}{\textit{Task-Spec.}}
& EEGNet
& 59.8 $\pm$ 1.5
& 51.0 $\pm$ 2.1
& 79.9 $\pm$ 0.4
& 42.4 $\pm$ 1.5
& 58.3 $\pm$ 0.5
& 23.4 $\pm$ 0.7
& 70.9 $\pm$ 0.8
\tabularnewline

&
& EEGConformer
& 58.8 $\pm$ 2.1
& 61.4 $\pm$ 7.9
& 79.0 $\pm$ 0.4
& 46.6 $\pm$ 3.3
& 57.0 $\pm$ 1.1
& 29.0 $\pm$ 0.5
& 72.3 $\pm$ 1.6
\tabularnewline

\cmidrule(lr){2-10}

&
\multirow{5}{*}{\textit{Foundation.}}
& BIOT
& 42.2 $\pm$ 5.9
& 63.7 $\pm$ 4.3
& 75.7 $\pm$ 0.2
& 39.2 $\pm$ 4.2
& 30.8 $\pm$ 2.2
& 11.2 $\pm$ 0.3
& 71.5 $\pm$ 1.1
\tabularnewline

&
& LaBraM
& 64.5 $\pm$ 3.8
& 67.2 $\pm$ 5.9
& 77.2 $\pm$ 1.5
& 52.5 $\pm$ 1.4
& 59.3 $\pm$ 0.7
& 26.8 $\pm$ 0.9
& 75.3 $\pm$ 0.9
\tabularnewline

&
& CBraMod
& 61.3 $\pm$ 0.8
& 70.7 $\pm$ 1.8
& \underline{80.9 $\pm$ 0.3}
& 52.0 $\pm$ 1.0
& 56.6 $\pm$ 1.2
& 29.4 $\pm$ 1.5
& 76.0 $\pm$ 0.9
\tabularnewline

&
& REVE
& \underline{67.4 $\pm$ 0.8}
& \underline{73.3 $\pm$ 3.8}
& 78.1 $\pm$ 1.5
& \underline{55.3 $\pm$ 3.5}
& \textbf{69.4 $\pm$ 0.9}
& \underline{32.0 $\pm$ 1.2}
& \underline{76.5 $\pm$ 0.8}
\tabularnewline

&
& LUNA
& 67.0 $\pm$ 1.6
& 68.8 $\pm$ 2.4
& 80.2 $\pm$ 0.6
& 49.7 $\pm$ 2.8
& 61.7 $\pm$ 0.6
& 29.3 $\pm$ 11.8
& 74.8 $\pm$ 0.8
\tabularnewline

\midrule

\textit{Multi-Task}
& \textit{Latent-Pred.}
& \textbf{BLPM (Ours)}
& \textbf{69.1 $\pm$ 1.4}
& \textbf{75.4 $\pm$ 2.1}
& \textbf{82.3 $\pm$ 0.2}
& \textbf{57.2 $\pm$ 1.1}
& \underline{68.2 $\pm$ 0.7}
& \textbf{34.3 $\pm$ 0.6}
& \textbf{77.1 $\pm$ 0.6}
\tabularnewline

\bottomrule
\end{tabular*}
}

\caption{Balanced Accuracy (B-Acc, \%) comparison across seven downstream datasets. The best and second-best results are marked in bold and underlined, respectively.}
\label{tab:bacc_comparison}
\end{table*}

\begin{table*}[!t]
\centering

\small
\renewcommand{\arraystretch}{1.2}
\setlength{\tabcolsep}{1.9pt}

\begin{tabular}{c c l c c c c c c c c c}
\toprule

\multirow{2}{*}{\textbf{Task}}
& \multirow{2}{*}{\textbf{Paradigm}}
& \multirow{2}{*}{\textbf{Method}}
& \multicolumn{3}{c}{\textbf{COG-BCI}}
& \multicolumn{3}{c}{\textbf{Mental Arithmetic}}
& \multicolumn{3}{c}{\textbf{TUEV}}
\\

\cmidrule(lr){4-6}
\cmidrule(lr){7-9}
\cmidrule(lr){10-12}

&
&
&
\textbf{B-Acc}
& \textbf{Kappa}
& \textbf{F1-W}
& \textbf{B-Acc}
& \textbf{AUPRC}
& \textbf{AUROC}
& \textbf{B-Acc}
& \textbf{Kappa}
& \textbf{F1-W}
\\

\midrule

\multirow{7}{*}{\textit{Single-Task}}
& \multirow{2}{*}{\textit{Task-Spec.}}
& EEGNet
& 59.8$\pm$1.5
& 39.6$\pm$2.3
& 58.9$\pm$1.4
& 51.0$\pm$2.1
& 50.5$\pm$2.6
& 50.3$\pm$4.3
& 42.4$\pm$1.5
& 27.1$\pm$2.2
& 39.1$\pm$2.6
\\

&
& EEGConformer
& 58.8$\pm$2.1
& 38.2$\pm$3.1
& 51.2$\pm$2.4
& 61.4$\pm$7.9
& 63.2$\pm$9.2
& 64.6$\pm$12.3
& 46.6$\pm$3.3
& 32.3$\pm$2.0
& 41.4$\pm$1.1
\\

\cmidrule(lr){2-12}

&
\multirow{5}{*}{\textit{Foundation.}}
& BIOT
& 42.2$\pm$5.9
& 13.4$\pm$8.9
& 38.1$\pm$10.7
& 63.7$\pm$4.3
& 65.6$\pm$4.5
& 70.9$\pm$4.6
& 39.2$\pm$4.2
& 29.1$\pm$5.2
& 42.6$\pm$5.3
\\

&
& LaBraM
& 64.5$\pm$3.8
& 46.7$\pm$5.8
& 61.2$\pm$6.2
& 67.2$\pm$5.9
& 77.4$\pm$4.4
& 81.4$\pm$3.7
& 52.5$\pm$1.4
& 36.6$\pm$4.2
& 48.9$\pm$4.2
\\

&
& CBraMod
& 61.3$\pm$0.8
& 41.9$\pm$1.2
& 59.7$\pm$1.7
& 70.7$\pm$1.8
& 72.7$\pm$1.6
& 78.1$\pm$2.3
& 52.0$\pm$1.0
& 39.2$\pm$3.5
& 50.8$\pm$3.6
\\

&
& REVE
& \underline{67.4$\pm$0.8}
& \underline{51.1$\pm$1.2}
& \underline{67.3$\pm$1.1}
& \underline{73.3$\pm$3.8}
& \underline{84.6$\pm$1.7}
& \underline{85.3$\pm$2.2}
& \underline{55.3$\pm$3.5}
& \underline{48.3$\pm$3.5}
& \underline{57.7$\pm$2.9}
\\

&
& LUNA
& 67.0$\pm$1.6
& 50.5$\pm$2.4
& 65.8$\pm$2.1
& 68.8$\pm$2.4
& 69.8$\pm$3.1
& 76.1$\pm$2.9
& 49.7$\pm$2.8
& 35.2$\pm$4.7
& 47.9$\pm$3.3
\\

\midrule

\textit{Multi-Task}
& \textit{Latent-Pred.}
& \textbf{BLPM (Ours)}
& \textbf{69.1$\pm$1.4}
& \textbf{52.3$\pm$1.1}
& \textbf{68.2$\pm$1.2}
& \textbf{75.4$\pm$2.1}
& \textbf{85.2$\pm$1.4}
& \textbf{86.1$\pm$1.7}
& \textbf{57.2$\pm$1.1}
& \textbf{50.1$\pm$1.8}
& \textbf{59.3$\pm$1.5}
\\

\bottomrule
\end{tabular}

\caption{Performance comparison on the COG-BCI, Mental Arithmetic, and TUEV datasets across multiple evaluation metrics. The best and second-best results are marked in bold and underlined, respectively.}
\label{tab:detailed_comparison2}
\end{table*}

\subsection{Implementation Details}
\subsubsection{Data Preprocessing.}

For pretraining, we apply a standardized preprocessing pipeline \cite{wang2025cbramod} with several modifications. Recordings shorter than five minutes are removed, the first and last minutes of each recording are discarded, and standard channels from the international 10-20 system are retained. The signals are then band-pass filtered at 0.3-75 Hz, notch filtered at 60 Hz, resampled to 200 Hz, and split into non-overlapping 30-second segments. Segments with absolute amplitudes exceeding 100 $\mu$V are excluded. The resulting EEG segments are used for self-supervised latent predictive pretraining.
For each task dataset, we use NeuralSet \cite{king2026neuralset} and NeuralBench \cite{banville2026neuralbench} to reproduce the preprocessing pipelines originally used for each baseline model. Since pretrained models depend on specific input distributions and signal statistics, we use model-specific preprocessing.

\subsubsection{Baseline Models.}
We compare BLPM against seven representative EEG decoding models under the same standardized evaluation framework. The baseline models are grouped into two categories based on their training strategies. The first category includes task-specific EEG decoding models, namely EEGNet \cite{lawhern2018eegnet} and EEGConformer \cite{song2023eegconformer}. EEGNet is a compact convolutional model for EEG classification, and EEGConformer integrates local convolutional encoding with Transformer-based global dependency modeling. The second category consists of EEG foundation models, including BIOT \cite{yang2023biot}, LaBraM \cite{jiang2024labram}, CBraMod \cite{wang2025cbramod}, REVE \cite{elouahidi2025reve}, and LUNA \cite{doner2025luna}. These models learn universal EEG representations through large-scale pretraining and are adapted separately to each downstream dataset. In contrast, BLPM is trained under a unified multi-task setting and performs instruction-conditioned semantic embedding prediction across all seven downstream tasks.

\subsubsection{Evaluation Metrics.}
We assess performance using different evaluation metrics for multiclass and binary classification settings. For multiclass classification, we report Balanced Accuracy, Cohen’s Kappa, and Weighted F1-score. For binary classification, we use Balanced Accuracy, AUPRC, and AUROC.

\subsubsection{Training \& Environment Settings.}
All experiments were conducted using four NVIDIA GeForce RTX 3090 GPUs with Python 3.12.13 and PyTorch 2.6. The best-performing checkpoint for each model was selected based on validation-set performance and subsequently evaluated on the test set. We report the mean and standard deviation over five independent runs with different seeds.

\subsubsection{LLM Backbone.}
Unlike prior studies that adopt the GPT-2 \cite{jiang2025neurolm,yang2025thdbar} or the Qwen 2.5 series \cite{wang2026kastbar}, we employ Llama 3.2-1B-Instruct \cite{dubey2024llama3} as our language model backbone. We adapt the model using low-rank adaptation (LoRA) \cite{hu2022lora} applied to the linear projection layers, enabling parameter-efficient adaptation.

\section{Experimental Results}

\subsection{Performance on Downstream Tasks}

To provide a comprehensive evaluation of downstream performance, we conduct experiments on seven representative tasks. The quantitative balanced accuracy results for all downstream tasks are summarized in Table~\ref{tab:bacc_comparison}. Table~\ref{tab:detailed_comparison2} reports results for COG-BCI, Mental Arithmetic, and TUEV across multiple evaluation metrics.

\subsubsection{Performance Comparison with Baselines.}

As shown in Table~\ref{tab:bacc_comparison} and Table~\ref{tab:detailed_comparison2}, BLPM demonstrates strong generalization capabilities across heterogeneous EEG decoding tasks.

\textbf{Comparison with Task-Specific Models:}
BLPM consistently outperforms the task-specific baselines, EEGNet and EEGConformer, across all seven downstream tasks, with substantial performance gains on most datasets, even though these baselines are trained specifically for each task. In particular, BLPM achieves performance improvements of \(14.0\%\), \(10.6\%\), and \(9.9\%\) over the best-performing task-specific baseline on Mental Arithmetic, TUEV, and PhysioNet-MI, respectively. These substantial performance gains indicate that BLPM generalizes more effectively across diverse downstream tasks with heterogeneous signal characteristics than task-specific models.

\textbf{Comparison with EEG Foundation Models:}
BLPM also demonstrates stronger performance than existing foundation models, which have shown robust performance across various downstream tasks. Compared with the strongest EEG foundation model baseline on each dataset, BLPM achieves a higher balanced accuracy on six datasets. Specifically, BLPM achieves a balanced accuracy of \(34.3\%\) on FACED, \(75.4\%\) on Mental Arithmetic, and \(57.2\%\) on TUEV, outperforming REVE, the strongest EEG foundation model baseline, by \(2.3\%\), \(2.1\%\), and \(1.9\%\), respectively. BLPM also increases performance on COG-BCI, TUAB, and HMC by \(1.7\%\), \(1.4\%\), and \(0.6\%\). These results show that BLPM consistently provides robust performance across nearly all downstream tasks spanning diverse EEG paradigms. On PhysioNet-MI, BLPM achieves the second-best performance with a balanced accuracy of \(68.2\%\), trailing REVE by only \(1.2\%\), while outperforming the next strongest foundation model baseline, LUNA by \(6.5\%\). Remarkably, BLPM achieves these improvements using a unified multi-task model, whereas the competing foundation models are separately fine-tuned on individual datasets. These overall performance gains indicate that continuous latent prediction and language-guided semantic decomposition learn transferable representations and effectively align them with task-relevant semantics across heterogeneous neural decoding tasks. Furthermore, these results suggest that representing discrete texts as continuous semantic representations in a shared embedding space and aligning them with EEG representations enables the model to generalize across datasets without separately optimized task-specific prediction heads.

\subsection{Ablation Studies}
To systematically evaluate the contributions of the key architectural components and training strategies in BLPM, we conducted comprehensive ablation studies on the TUEV, FACED, and HMC datasets. The detailed results are presented in Table~\ref{tab:ablation_study}.

\subsubsection{Effect of the Pretraining Objective.}
To examine the effectiveness of latent prediction for EEG pretraining, we compare embedding-space prediction with input-space reconstruction by replacing our latent prediction objective with a masked reconstruction objective. As shown in Table~\ref{tab:ablation_study}(a), the masked reconstruction variant achieves balanced accuracies of 54.63\%, 32.71\%, and 76.84\% on TUEV, FACED, and HMC, respectively. In contrast, our model achieves superior performance across all three datasets, with balanced accuracies of 57.18\%, 34.29\%, and 77.10\%, corresponding to improvements of up to 2.55\%p. These results demonstrate that latent representation prediction captures task-relevant EEG dynamics more effectively than masked autoencoding-based approaches, which are more susceptible to noisy, low-level signal variations.

\subsubsection{Effect of the EEG-Language Semantic Alignment.}
To validate the effect of EEG-language semantic alignment, we remove the alignment stage between EEG representations and language-derived semantic embeddings while retaining the remaining components of the framework. As shown in Table~\ref{tab:ablation_study}(b), eliminating the alignment stage reduces performance from 57.18 to 52.02 on TUEV, from 34.29 to 27.38 on FACED and from 77.10 to 72.17 on HMC, with the largest decrease of 6.91\% observed on FACED. These results demonstrate that the alignment stage plays an important role in bridging the modality gap between EEG and language representations, thereby facilitating the integration of continuous neural signals and textual semantic information.

\subsubsection{Effect of the Multi-Query Semantic Decomposition.}
We compare the proposed MQSD module with two simplified alternatives. As shown in Table~\ref{tab:ablation_study}(c), the global-pooling variant directly averages the projected EEG tokens without semantic queries. In contrast, Table~\ref{tab:ablation_study}(d) uses randomly initialized learnable queries with the same query-based aggregation structure as MQSD, but without language-derived semantic guidance. The global-pooling variant achieves balanced accuracies of 55.61\%, 32.54\%, and 75.02\%, while the learnable-query variant achieves 56.36\%, 33.48\%, and 76.58\% on TUEV, FACED, and HMC, respectively. The full model with the MQSD module in Table~\ref{tab:ablation_study}(f) further improves performance to 57.18\%, 34.29\%, and 77.10\%. These results demonstrate that MQSD decomposes EEG representations into multiple semantically guided components, enabling the model to capture complementary neural patterns that are not fully preserved by global pooling or generic learnable queries.

\subsubsection{Effect of the Semantic Embedding Prediction.}
To evaluate the effect of the semantic prediction target, we replace continuous semantic embedding prediction with autoregressive next-token prediction in the token space while retaining the remaining components of the framework. Specifically, the autoregressive variant predicts the class-specific answer token according to its conditional likelihood, whereas our approach predicts a continuous semantic embedding corresponding to the target class. As shown in Table~\ref{tab:ablation_study}(e), the token-prediction variant achieves balanced accuracies of 55.27\%, 32.92\%, and 75.96\% on TUEV, FACED, and HMC, respectively. In comparison, the full BLPM in Table~\ref{tab:ablation_study}(f) achieves 57.18\%, 34.29\%, and 77.10\%, corresponding to improvements of 1.91\%, 1.37\%, and 1.14\%, respectively. These results indicate that predicting continuous semantic embeddings provides a more effective supervision signal for EEG decoding tasks than generating discrete answer tokens based on conditional likelihood, as it focuses on aligning EEG representations with natural-language semantics rather than optimizing the generation probabilities of token forms.

\begin{table}[t]
\centering
\footnotesize
\setlength{\tabcolsep}{2.6pt}
\renewcommand{\arraystretch}{1.12}

\begin{tabular*}{\columnwidth}{
@{\extracolsep{\fill}}
c c c c c c c c
@{}
}
\toprule
\multirow{2}{*}{\textbf{ID}}
& \multicolumn{4}{c}{\textbf{Model Configuration}}
& \multicolumn{3}{c}{\textbf{B-Acc (\%)}}
\\
\cmidrule(lr){2-5}
\cmidrule(lr){6-8}
&
\textbf{Pretrain}
& \textbf{Align.}
& \textbf{Query}
& \textbf{Pred.}
& \textbf{TUEV}
& \textbf{FACED}
& \textbf{HMC}
\\
\midrule

(a)
& Recon.
& \checkmark
& MQSD
& Embed.
& 54.63 & 32.71 & 76.84
\\

(b)
& CELP
& $\times$
& MQSD
& Embed.
& 52.02 & 27.38 & 72.17
\\

(c)
& CELP
& \checkmark
& GP
& Embed.
& 55.61 & 32.54 & 75.02
\\

(d)
& CELP
& \checkmark
& LQ
& Embed.
& 56.36 & 33.48 & 76.58
\\

(e)
& CELP
& \checkmark
& MQSD
& Token
& 55.27 & 32.92 & 75.96
\\

(f)
& CELP
& \checkmark
& MQSD
& Embed.
& \textbf{57.18}
& \textbf{34.29}
& \textbf{77.10}
\\
\bottomrule
\end{tabular*}

\caption{Ablation study on TUEV, FACED, and HMC. Rows (a)--(e) each differ from the full model (f) in a single component: \textbf{Pretrain}, the pretraining objective (Recon., masked reconstruction; CELP, our latent prediction); \textbf{Align.}, the EEG-language semantic alignment stage; \textbf{Query}, the query mechanism (GP, global pooling; LQ, generic learnable queries; MQSD, our multi-query semantic decomposition); \textbf{Pred.}, the prediction target (Token, autoregressive token generation; Embed., semantic embedding prediction).}
\label{tab:ablation_study}
\end{table}
\section{Conclusion}
In this paper, we introduce BLPM, a continuous latent predictive EEG–language foundation model that reformulates heterogeneous EEG decoding as a semantic embedding prediction problem. By replacing discrete tokenization and autoregressive modeling with continuous semantic prediction, BLPM aligns EEG representations with language semantics in a continuous latent space rather than a discrete token space. Across diverse EEG benchmarks, BLPM demonstrates strong performance on heterogeneous neural decoding tasks. These results support continuous latent prediction as an effective modeling paradigm for universal EEG decoding.




\bibliography{references}

\clearpage
\urlstyle{rm} 
\def\UrlFont{\rm}  
\frenchspacing  

\pdfinfo{
/TemplateVersion (2027.1)
}

\setcounter{secnumdepth}{1}
\graphicspath{{figures/}}




\appendix

\setcounter{table}{0}
\renewcommand{\thetable}{S\arabic{table}}

\setcounter{figure}{0}
\renewcommand{\thefigure}{S\arabic{figure}}

\section{Related Work}
\subsection{Conventional EEG Decoding Methods}
Conventional EEG decoding methods predominantly relied on supervised learning to extract task-relevant features from labeled EEG data. EEGNet~\cite{lawhern2018eegnet} introduced a compact convolutional architecture employing depthwise and separable convolutions to efficiently capture temporal and spatial EEG features. EEGConformer~\cite{song2023eegconformer} combined convolutional feature extraction with transformer-based self-attention to jointly model local patterns and long-range temporal dependencies. Despite their effectiveness, these models are generally trained for specific tasks and datasets and depend heavily on labeled data, while much of the available EEG data remains unlabeled. These limitations have motivated the development of EEG foundation models that learn generalizable representations from large-scale unlabeled EEG data.

\subsection{EEG Foundation Models}
Early EEG foundation models sought to learn transferable representations from heterogeneous EEG datasets with varying electrode configurations, sampling rates, and signal lengths. BIOT~\cite{yang2023biot}, for example, enabled cross-dataset pretraining through channel-wise tokenization that accommodated diverse montages and sampling rates. Subsequent studies have primarily adopted three approaches: masked autoencoding, discrete neural tokenization, and autoregressive modeling. These approaches are not mutually exclusive and have often been combined within a single framework.

\subsubsection{Masked Autoencoding.}
Universal EEG representations are commonly learned by masking portions of the input and reconstructing raw EEG signals, latent representations, or discrete tokens. EEGPT~\cite{wang2024eegpt} employed a dual self-supervised pretraining objective that combined masked EEG reconstruction with spatio-temporal representation alignment to learn robust and generalizable EEG representations. CBraMod~\cite{wang2025cbramod} proposed a criss-cross transformer designed to separately model spatial and temporal dependencies among EEG patches through parallel attention mechanisms and pretrained it using masked EEG reconstruction. CSBrain~\cite{zhou2026csbrain} introduced cross-scale spatio-temporal tokenization to aggregate multi-scale neural patterns within temporal windows and anatomical brain regions, along with structured sparse attention to model long-range dependencies across windows and regions. It was pretrained by reconstructing masked EEG segments. REVE~\cite{elouahidi2025reve} introduced a 4D spatio-temporal positional encoding scheme to accommodate arbitrary electrode configurations and signal lengths. It was pretrained using spatio-temporal block masking and masked raw EEG reconstruction on a large-scale heterogeneous EEG corpus.

\subsubsection{Discrete Neural Tokenization.}
EEG signals can also be quantized into discrete neural tokens that provide compact representations of continuous EEG embeddings and serve as prediction targets during pretraining. LaBraM~\cite{jiang2024labram} introduced a vector-quantized neural tokenizer trained via Fourier-spectrum prediction to encode continuous EEG channel patches into discrete neural codes. NeuroLM~\cite{jiang2025neurolm} and THD-BAR~\cite{yang2025thdbar} similarly learned discrete EEG tokenizers and further combined them with autoregressive modeling. CodeBrain~\cite{ma2026codebrain} introduced a TFDual-Tokenizer that quantized joint time-frequency EEG embeddings using separate temporal and frequency codebooks, together with a multi-scale architecture that captured sparse long-range dependencies through structured global convolution and local intra-patch dependencies through sliding-window attention. During pretraining, the model predicted the temporal and frequency token indices of masked EEG patches.

\subsubsection{Autoregressive Modeling.}
Building on discrete neural tokenization, recent models learn causal dependencies among EEG tokens and integrate the token sequences into autoregressive generative models. NeuroLM~\cite{jiang2025neurolm} learned a text-aligned neural tokenizer through vector-quantized temporal-frequency prediction and integrated the resulting neural tokens into an LLM through multi-channel autoregressive pretraining and multi-task instruction tuning. THD-BAR~\cite{yang2025thdbar} introduced a brain topology hierarchy and developed a topology-hierarchical VQ-VAE for multi-scale discrete EEG tokenization, followed by autoregressive pretraining based on next-scale-time prediction. KAST-BAR~\cite{wang2026kastbar} combined topology-preserving discrete EEG tokenization with expert-level semantic profiles generated from physiological features and jointly pretrained textual knowledge and discrete EEG tokens using next-token prediction.

\subsection{Cross-Modal Alignment and LLM Integration}
Recent studies have extended brain-signal representation learning by aligning neural signals with semantic representations from linguistic, visual, and acoustic modalities. 
BrainMosaic~\cite{li2026assembling} introduced semantic intent decoding, which decomposed EEG and sEEG signals into compositional semantic units and reconstructed natural-language descriptions from the decoded intents. MindMix~\cite{liu2026mindmix} aligned EEG representations with neural-acoustic features to support transferable decoding across auditory perception tasks. Hierarchical visual alignment~\cite{zheng2026learning} mapped brain signals to multi-scale visual embeddings extracted from complementary pretrained visual encoders, preserving both high-level semantic information and fine-grained visual details.

Beyond cross-modal representation alignment, recent studies have integrated neural signals directly into pretrained language models or multimodal LLMs for generative interpretation. In the clinical domain, CerebraGloss~\cite{gu2026cerebragloss} treated EEG waveforms as visual inputs and instruction-tuned a pretrained vision-language model on paired EEG images and clinical text for fine-grained clinical EEG interpretation. CELM~\cite{pradeepkumar2026neural} connected a long-context EEG encoder with an LLM through sequence-aware alignment, enabling clinical report generation directly from long-duration EEG recordings. Collectively, these studies demonstrate the potential of cross-modal alignment and pretrained language models for semantically grounding neural representations. However, most existing approaches rely on modality-specific paired data or generative decoding, whereas BLPM aligns EEG representations with language through a shared semantic space using language-derived supervision without autoregressive text generation.

\begin{table*}[t!]
\centering
\setlength{\tabcolsep}{1.5pt}
\renewcommand{\arraystretch}{1.30}
\begin{tabular}{l l l c c c c c c}
\toprule
\textbf{Category} & \textbf{Task} & \textbf{Dataset} & \textbf{\#Channels} &
\textbf{Rate} & \textbf{Duration} & \textbf{\#Subjects} & \textbf{\#Samples} &
\textbf{\#Classes} \\
\midrule
\multirow{2}{*}{Clinical}
  & Pathology detection            & TUAB             & 16 & 256\,Hz & 5\,s  & 2{,}383 & 409k  & 2 \\
  & Clinical event detection       & TUEV             & 16 & 250\,Hz & 3\,s  & 370     & 112k  & 6 \\
\cmidrule(lr){1-9}
Sleep
  & Sleep stage classification     & HMC              & 4  & 256\,Hz & 30\,s & 151     & 137k  & 5 \\
\cmidrule(lr){1-9}
\multirow{3}{*}{Internal state}
  & Mental workload classification & COG-BCI          & 64 & 500\,Hz & 5\,s  & 29      & 15.4k & 3 \\
  & Mental stress detection        & Mental Arithmetic & 20 & 500\,Hz & 5\,s & 36      & 1.7k  & 2 \\
  & Emotion classification         & FACED            & 32 & 250\,Hz & 5\,s  & 123     & 10.3k & 9 \\
\cmidrule(lr){1-9}
BCI
  & Motor imagery classification   & PhysioNet-MI     & 64 & 160\,Hz & 4\,s  & 109     & 9.8k  & 4 \\
\bottomrule
\end{tabular}
\caption{The seven downstream datasets, grouped by \mbox{NeuralBench} task
category. \emph{\#Channels}: number of EEG electrodes; \emph{Rate}: native
sampling rate; \emph{Duration}: window length fed to the model;
\emph{\#Subjects}\,/\,\emph{\#Samples}: participant and labelled-segment counts
(rounded to thousands, k); \emph{\#Classes}: number of target labels.}
\label{tab:datasets}
\end{table*}

\section{Dataset details and Preprocessing}
\label{app:datasets}
We evaluate BLPM on seven downstream datasets spanning four
\mbox{NeuralBench}~\citep{banville2026neuralbench} task categories---\emph{clinical}, \emph{sleep}, \emph{internal-state}, and \emph{BCI}
decoding. Table~\ref{tab:datasets} summarizes their recording
configurations; they differ widely in electrode montage, sampling rate,
and scale, providing a heterogeneous evaluation setting for cross-task and cross-dataset generalization.

\begin{itemize}
\item \textbf{Clinical --- TUAB}~\citep{lopez2015automated}: binary
normal/abnormal EEG classification from the Temple University Hospital Abnormal corpus.

\item \textbf{Clinical --- TUEV}~\citep{harati2015improved}: six-class
EEG event-type classification from the Temple University Hospital Event
corpus.

\item \textbf{Sleep --- HMC}~\citep{PhysioNet-hmc-sleep-staging-1.1}:
five-class sleep-stage classification over 30-second epochs from the
Haaglanden Medisch Centrum sleep database.

\item \textbf{Internal state --- COG-BCI}~\citep{hinss2023open}:
three-class mental workload classification.

\item \textbf{Internal state --- Mental Arithmetic}~\citep{zyma2019workload}:
binary mental stress detection.

\item \textbf{Internal state --- FACED}~\citep{chen2023large}: nine-class emotion recognition using 32-channel EEG recordings elicited by affective video stimuli.

\item \textbf{BCI --- PhysioNet-MI}~\citep{PhysioNet-eegmmidb-1.0.0}:
four-class motor-imagery classification from 64-channel EEG.
\end{itemize}

All datasets are processed and evaluated using a unified pipeline implemented with NeuralSet~\cite{king2026neuralset}, including signal preprocessing, window extraction, and subject-level data partitioning.

The following paragraphs provide detailed descriptions of each dataset, including its participants, recording configuration, and task formulation.

\subsubsection{TUAB.}
TUAB is a large-scale clinical EEG dataset used for binary abnormal EEG detection~\citep{lopez2015automated}. Each clinical EEG recording is labeled as either normal or abnormal based on interpretations provided by clinical EEG experts. The task evaluates whether a model can identify pathological or otherwise abnormal patterns in routine clinical EEG recordings.

Following the preprocessing protocol used in our experiments, we retain 16 channels with most recordings sampled at 256\,Hz. The signals are divided into non-overlapping 5-s windows. We follow the official TUAB partition: recordings in the official evaluation subset are used exclusively as the test set. The remaining official training data are divided at the subject level, with 20\% of the training subjects assigned to validation using random state 33. This preserves the official test set while preventing recordings from the same patient from appearing in both training and validation subsets.

\subsubsection{TUEV.}
TUEV is a clinical EEG event dataset used to classify pathological events and common non-cerebral patterns in clinical EEG recordings~\citep{harati2015improved}. We use the standard six event categories: spike-and-sharp wave (SPSW), generalized periodic epileptiform discharge (GPED), periodic lateralized epileptiform discharge (PLED), eye movement (EYEM), artifact (ARTF), and background activity (BCKG). The first three categories represent clinically relevant epileptiform patterns, whereas the remaining categories represent eye movements, artifacts, and background EEG activity.

Following our experimental preprocessing, 16 EEG channels are retained and all signals are resampled to 120\,Hz. Event-centered EEG segments are extracted using 3-s windows with a 1-s stride, resulting in a 2-s overlap, and are assigned one of the six event labels. We use the official TUEV evaluation partition as the test set. The official development data are divided at the subject level into training and validation subsets, with 20\% of development subjects assigned to validation using random state 33. Consequently, no patient is shared across the training, validation, and test subsets.

\subsubsection{HMC.}
The Haaglanden Medisch Centrum (HMC) sleep staging database is a clinical polysomnography dataset used for automatic sleep-stage classification from whole-night recordings \citep{PhysioNet-hmc-sleep-staging-1.1}. It contains 151 recordings acquired from a heterogeneous clinical population referred for the assessment of sleep disorders.

All channels were resampled to 200\,Hz. Sleep stages were manually annotated in non-overlapping 30-s epochs according to standard clinical sleep-scoring guidelines. The original database provides five stage labels: Wake, N1, N2, N3, and REM. Recordings are divided at the subject level into training, validation, and test subsets, with validation and test ratios of 0.2 using random state 33. Prior to epoch extraction, long wake periods occurring before the first sleep epoch and after the last sleep epoch are limited to a maximum of 30 minutes.

\subsubsection{COG-BCI.}
COG-BCI is a multi-session cognitive EEG dataset developed for passive brain-computer interface research and cognitive-state decoding \citep{hinss2023open}. The full dataset contains recordings from 29 participants collected across three sessions while they performed four cognitive tasks: MATB-II, N-back, psychomotor vigilance, and Flanker tasks. In this study, we use the MATB-II recordings for three-class mental workload classification, corresponding to easy, medium, and difficult workload conditions.

The EEG signals were originally recorded from 64 channels at 500\,Hz and resampled to 120\,Hz. Following the NeuralBench workload protocol, only the \texttt{MATBeasy}, \texttt{MATBmed}, and \texttt{MATBdiff} conditions are retained. Each continuous task recording is divided into non-overlapping 5-s windows, and incomplete windows are discarded. Subjects are divided into disjoint training, validation, and test sets, with validation and test ratios of 0.2.

\subsubsection{Mental Arithmetic.}
The Mental Arithmetic dataset is a cognitive EEG dataset used to distinguish resting-state brain activity from activity recorded during mental arithmetic \citep{zyma2019workload}. It contains recordings from 36 healthy participants under two conditions: a resting background condition and a mental serial-subtraction condition.

The EEG signals were acquired from 20 scalp channels at 500\,Hz and resampled to 200\,Hz. During the original data collection, a 0.5-Hz high-pass filter, a 45-Hz low-pass filter, and a 50-Hz power-line notch filter were applied. The released data include artifact-free resting and mental-arithmetic recordings selected through expert visual inspection. Following NeuralBench \cite{banville2026neuralbench}, the signals are divided into non-overlapping 5-s windows. Subjects are assigned to mutually exclusive training, validation, and test subsets, with validation and test ratios of 0.2 using random state 33. Class weights are estimated from the training data to address class imbalance.

\subsubsection{FACED.}
FACED is an affective EEG dataset designed for fine-grained emotion recognition from neural responses elicited by emotional video stimuli \citep{chen2023large}. It contains EEG recordings from 123 participants who watched 28 video clips covering nine emotion categories: amusement, inspiration, joy, tenderness, anger, fear, disgust, sadness, and neutral emotion. EEG signals were recorded from 32 channels, with the original acquisition rate differing between the two data-collection cohorts (250 or 1000\,Hz), and were provided after downsampling to 250\,Hz.

Following the NeuralBench emotion-recognition formulation, only the last 30\,s of each video presentation are retained, and the selected signals are divided into non-overlapping 5-s windows, with up to six segments extracted per video clip. Each window is assigned the emotion category associated with the corresponding video clip. The dataset is divided into mutually exclusive training, validation, and test subjects using a subject-disjoint split. The validation and test ratios are both set to 0.2, using random state 33. Class weights are computed from the training subset to mitigate the substantial imbalance among emotion categories.

\subsubsection{PhysioNet-MI.}
PhysioNet-MI is a motor imagery EEG dataset used to classify imagined movements of different body parts \citep{PhysioNet-eegmmidb-1.0.0}. It contains 64-channel EEG recordings collected from 109 participants using the BCI2000 system. Participants performed both motor execution and motor imagery tasks involving the left fist, right fist, both fists, and both feet. In this work, only motor imagery trials are used, resulting in a four-class motor imagery classification task.

The EEG signals were acquired at 160\,Hz and resampled to 120\,Hz. Following the NeuralBench motor imagery protocol, only events whose annotations correspond to motor imagery are retained, while motor execution events are excluded. A single 4-s EEG segment is extracted from each imagery trial onset without sliding-window overlap. Subjects S088, S092, and S100 are excluded according to the NeuralBench dataset configuration. Because PhysioNet-MI does not provide an official machine-learning train/test partition, the remaining subjects are divided into subject-disjoint training, validation, and test sets. The validation and test ratios are both 0.2, using random state 33.

\section{Baseline Models}
\label{app:baseline_models}
We compare BLPM against two widely adopted supervised models and five recent foundation models for EEG representation learning. For all pretrained baselines, we initialize the encoder using the publicly released pretrained checkpoint and adapt the model to each downstream dataset with a task-specific classification head. The task-specific models are trained from scratch under the same downstream evaluation protocol.
\subsection{Task-specific Models}

\subsubsection{EEGNet.}
EEGNet~\cite{lawhern2018eegnet} is a compact convolutional neural network architecture designed to generalize across different EEG-based brain-computer interface paradigms. It first applies temporal convolutions to learn frequency-selective filters and then uses depthwise spatial convolutions to capture channel-specific spatial patterns. A subsequent separable convolution combines the learned temporal and spatial features while maintaining a small number of trainable parameters. Given its computational efficiency and broad applicability, we include EEGNet as a representative lightweight task-specific baseline.

\subsubsection{EEGConformer.}
EEGConformer~\cite{song2023eegconformer} combines convolutional feature extraction with transformer-based sequence modeling. Its convolution module uses temporal and spatial convolutions to obtain local EEG features and partitions the resulting feature map into temporal tokens. These tokens are then processed by multi-head self-attention layers to model long-range dependencies across time. A fully connected classification head maps the resulting representation to the downstream label space. We train EEGConformer separately on each target dataset as a supervised baseline.

\subsection{Foundation Models}

\subsubsection{BIOT.}
BIOT~\cite{yang2023biot} is a linear transformer-based model developed for cross-dataset learning from heterogeneous biosignals. To accommodate differences in channel configurations, sequence lengths, and missing observations, BIOT independently divides each signal channel into fixed-duration segments and rearranges the resulting segments into a unified token sequence. Channel and relative positional embeddings retain the spatial
and temporal identity of individual tokens, while a linear transformer models their interactions with reduced computational complexity. We use its pretrained biosignal encoder and optimize a task-specific classification head for each downstream task.

\subsubsection{LaBraM.}

LaBraM~\cite{jiang2024labram} is a large-scale EEG foundation model trained using discrete neural tokenization and masked EEG modeling. Raw EEG recordings are divided into channel-wise temporal patches, and a vector-quantized neural tokenizer maps each continuous patch to a discrete neural code. The transformer encoder is subsequently pretrained to recover the neural codes of masked patches from the visible context. Temporal and spatial embeddings allow the model to process recordings with different durations and electrode configurations. For downstream evaluation, we fine-tune the pretrained LaBraM encoder together with a task-specific classification head.

\subsubsection{CBraMod.}

CBraMod~\cite{wang2025cbramod} introduces a criss-cross transformer for learning structured spatio-temporal EEG representations. Instead of flattening all channel and temporal patches into a single sequence, its transformer blocks apply spatial and temporal attention in parallel, explicitly accounting for the heterogeneous dependencies along the two dimensions. CBraMod further uses asymmetric conditional positional encoding to generate input-dependent positional information for EEG recordings with varying channel arrangements. The model is pretrained on the Temple University Hospital EEG Corpus using masked patch reconstruction and is subsequently fine-tuned for individual downstream tasks.

\subsubsection{REVE.}

REVE~\cite{elouahidi2025reve} is an EEG foundation model designed to transfer across heterogeneous recording systems and electrode layouts. Each EEG channel is divided into temporal patches, which are augmented with a four-dimensional spatio-temporal positional encoding derived from the three-dimensional electrode coordinates and temporal positions. During pretraining, contiguous regions are masked across the spatial and temporal dimensions, and a decoder reconstructs the missing raw EEG segments. REVE is pretrained on more than 60,000 hours of EEG collected from 92 datasets and approximately 25,000 participants. We adapt the released pretrained encoder to each target dataset using the corresponding downstream classification head.

\subsubsection{LUNA.}
LUNA~\cite{doner2025luna} is a topology-agnostic EEG foundation model that maps variable electrode configurations into a fixed-size latent representation. Learned latent queries attend to the available channel features through cross-attention, thereby compressing an arbitrary number of electrodes into a common latent space. Temporal transformer blocks operate on this unified representation rather than directly on all channel-time pairs, making the computational cost scale linearly with the number of input channels. LUNA is pretrained on more than 21,000 hours of EEG using masked patch reconstruction. For downstream tasks, we fine-tune the pretrained encoder and an attached task-specific prediction head.

\section{Experimental Settings}
\label{app:experimental_settings}

BLPM is trained in three stages. In Stage 1, the CELP encoder is pretrained using a 12-layer transformer with a hidden dimension of 384 and six attention heads. The predictor consists of six transformer layers with the same hidden dimension. Pretraining uses Smooth L1 loss with multiblock masking, where the context-mask scale ranges from 0.75 to 1.0, the prediction-mask scale ranges from 0.15 to 0.30, and the target union ratio ranges from 0.40 to 0.60. The model is optimized with AdamW using a batch size of 128, a peak learning rate of $5\times10^{-5}$, and a cosine schedule with five warmup epochs. The momentum coefficient of the target encoder is gradually increased from 0.996 to 1.0.

In Stage 2, semantic alignment is performed with task- and class-balanced sampling. We use a batch size of 64, AdamW optimization, a peak learning rate of $1\times10^{-4}$, and a warmup-cosine learning rate schedule. Training is conducted for 20 epochs with a temperature of 0.07 and gradient clipping of 1.0.

In Stage 3, downstream multi-task instruction tuning is performed using task- and class-balanced sampling with a batch size of 64. The model is optimized for five epochs using AdamW with a peak learning rate of $5\times10^{-4}$ and a minimum learning rate of $5\times10^{-5}$. LoRA is applied to the \texttt{q\_proj} and \texttt{v\_proj} modules with rank 8, scaling factor 16, and dropout 0.05.

Tables~\ref{tab:celp_hyperparameters}--\ref{tab:instruction_tuning_hyperparameters}
provide the complete hyperparameter settings for the three training stages.

\begin{table}[t]
\centering
\begin{tabular}{lc}
\toprule
\textbf{Hyperparameters} & \textbf{CELP Encoder} \\
\midrule
Encoder dimension        & 384 \\
Encoder depth            & 12 \\
Attention heads          & 6 \\
MLP ratio                & 4 \\
Dropout                  & 0.1 \\
Predictor dimension      & 384 \\
Predictor depth          & 6 \\
Pretraining loss         & Smooth L1 \\
Masking strategy         & Multiblock \\
Context block scale      & 0.75--1.0 \\
Prediction block scale   & 0.15--0.30 \\
Target union ratio       & 0.40--0.60 \\
EMA momentum             & 0.996 $\rightarrow$ 1.0 \\
Batch size               & 128 \\
Peak learning rate       & $5\times10^{-5}$ \\
Minimum learning rate    & $1\times10^{-5}$ \\
Learning rate scheduler  & Warmup + cosine \\
Optimizer                & AdamW \\
Adam $\beta$             & $(0.9, 0.95)$ \\
Weight decay             & 0.05 \\
Total epochs             & 30 \\
Warmup epochs            & 5 \\
Gradient clipping        & 0.5 \\
\bottomrule
\end{tabular}
\caption{Hyperparameters for CELP encoder pretraining.}
\label{tab:celp_hyperparameters}
\end{table}

\begin{table}[t]
\centering
\begin{tabular}{lc}
\toprule
\textbf{Hyperparameters} & \textbf{Semantic Alignment} \\
\midrule
Task sampling             & Yes \\
Class sampling            & Yes \\
Batch size                & 64 \\
Peak learning rate        & $1\times10^{-4}$ \\
Learning rate scheduler   & Warmup + cosine \\
Optimizer                 & AdamW \\
Adam $\beta$              & $(0.9, 0.95)$ \\
Weight decay              & 0.05 \\
Total epochs              & 20 \\
Warmup ratio              & 0.05 \\
Gradient clipping         & 1.0 \\
Temperature               & 0.07 \\
\bottomrule
\end{tabular}
\caption{Hyperparameters for semantic alignment.}
\label{tab:alignment_hyperparameters}
\end{table}

\begin{table}[t]
\centering
\begin{tabular}{lc}
\toprule
\textbf{Hyperparameters} & \textbf{Instruction Tuning} \\
\midrule
Task sampling               & Yes \\
Class sampling              & Yes \\
Batch size                  & 64 \\
Peak learning rate          & $5\times10^{-4}$ \\
Minimum learning rate       & $5\times10^{-5}$ \\
Learning rate scheduler     & Cosine \\
Optimizer                   & AdamW \\
Adam $\beta$                & $(0.9, 0.95)$ \\
Weight decay                & 0.1 \\
Total epochs                & 5 \\
Warmup ratio                & 0.1 \\
Gradient clipping           & 1.0 \\
LoRA rank / alpha / dropout & 8 / 16 / 0.05 \\
LoRA target modules         & \texttt{q\_proj}, \texttt{v\_proj} \\
\bottomrule
\end{tabular}
\caption{Hyperparameters for downstream multi-task instruction tuning.}
\label{tab:instruction_tuning_hyperparameters}
\end{table}

\section{Evaluation Metrics}
We evaluate binary classification using balanced accuracy, AUPRC, and AUROC, and multi-class classification using balanced accuracy, Cohen's \(\kappa\), and the weighted F1-score.

\subsection{Metrics for Binary Classification}

\subsubsection{Balanced Accuracy (B-Acc).}
Balanced accuracy assigns equal importance to the positive and negative classes and is therefore suitable for evaluating classification performance under class imbalance. It is defined as the average of the true-positive rate and the true-negative rate:
\begin{equation}
\text{B-Acc}
=
\frac{1}{2}
\left(
\frac{\text{TP}}{\text{TP}+\text{FN}}
+
\frac{\text{TN}}{\text{TN}+\text{FP}}
\right),
\label{eq:binary_balanced_accuracy}
\end{equation}
where \(\text{TP}\), \(\text{TN}\), \(\text{FP}\), and
\(\text{FN}\) denote the numbers of true-positive, true-negative, false-positive, and false-negative predictions, respectively. Unlike conventional accuracy, balanced accuracy is less affected by skewed class distributions.

\subsubsection{Area Under the Precision-Recall Curve (AUPRC).}
The precision-recall curve characterizes the trade-off between precision and recall as the classification threshold varies. Precision and recall are defined as
\begin{equation}
\text{Precision}
=
\frac{\text{TP}}{\text{TP}+\text{FP}},
\qquad
\text{Recall}
=
\frac{\text{TP}}{\text{TP}+\text{FN}}.
\label{eq:precision_recall}
\end{equation}
AUPRC summarizes the area under this curve and evaluates the model's ability to identify positive samples while limiting false-positive predictions across different thresholds. Because precision and recall do not explicitly depend on the number of true-negative samples, AUPRC is particularly informative for imbalanced classification problems in which the positive class is rare.

\subsubsection{Area Under the Receiver Operating Characteristic Curve (AUROC).}
The receiver operating characteristic curve represents the relationship between the true-positive rate and the false-positive rate as the classification threshold varies. These quantities are defined as
\begin{equation}
\text{TPR}
=
\frac{\text{TP}}{\text{TP}+\text{FN}},
\qquad
\text{FPR}
=
\frac{\text{FP}}{\text{TN}+\text{FP}}.
\label{eq:tpr_fpr}
\end{equation}
AUROC summarizes the area under this curve and measures the model's ability to discriminate between positive and negative samples across all classification thresholds. An AUROC of \(1.0\) indicates perfect discrimination, whereas an AUROC of \(0.5\) corresponds to chance-level discrimination.

\subsection{Metrics for Multi-class Classification}
\subsubsection{Balanced Accuracy (B-Acc).}
Balanced accuracy evaluates classification performance by assigning equal importance to all classes, regardless of their sample frequencies.
For a multi-class classification problem with \(C\) classes, it is computed as the macro-average of the class-wise recall:
\begin{equation}
\text{B-Acc}
=
\frac{1}{C}
\sum_{c=1}^{C}
\frac{\text{TP}_{c}}
{\text{TP}_{c}+\text{FN}_{c}},
\label{eq:multiclass_balanced_accuracy}
\end{equation}
where \(\text{TP}_{c}\) and \(\text{FN}_{c}\) denote the numbers of true-positive and false-negative predictions for class \(c\), respectively. Unlike conventional accuracy, balanced accuracy is less affected by class imbalance because each class contributes equally to the final score.

\subsubsection{Cohen's Kappa Coefficient (\(\kappa\)).}
Cohen's kappa coefficient measures the agreement between predicted and ground-truth labels while accounting for the agreement expected by chance. It is defined as
\begin{equation}
\kappa
=
\frac{p_o - p_e}{1 - p_e},
\label{eq:cohens_kappa}
\end{equation}
where \(p_o\) denotes the observed agreement and \(p_e\) denotes the expected agreement by chance based on the marginal label distributions.
A value of \(\kappa = 1\) indicates perfect agreement, whereas \(\kappa = 0\) corresponds to chance-level agreement. Negative values indicate agreement below the level expected by chance.

\subsubsection{Weighted F1-Score (F1-W).}
The F1-score for class \(c\) is defined as the harmonic mean of its precision and recall:
\begin{equation}
\text{F1}_{c}
=
\frac{
2\,\text{Precision}_{c}\,\text{Recall}_{c}
}{
\text{Precision}_{c}+\text{Recall}_{c}
}.
\label{eq:classwise_f1}
\end{equation}
The weighted F1-score is then computed by averaging the class-wise F1-scores according to the number of samples in each class:
\begin{equation}
\text{F1-W}
=
\sum_{c=1}^{C}
\frac{n_c}{N}
\text{F1}_{c},
\label{eq:weighted_f1}
\end{equation}
where \(n_c\) denotes the number of ground-truth samples belonging to class \(c\), and \(N=\sum_{c=1}^{C} n_c\) denotes the total number of samples. This weighting gives greater influence to classes with larger sample sizes.

\section{Algorithms}
\label{app:algorithms}

BLPM relies on self-supervised pretraining of the encoder
(Algorithm~\ref{alg:celp}) and instruction-conditioned inference via
semantic matching (Algorithm~\ref{alg:infer}). We detail the design of
each component below.

\begin{algorithm}[!t]
\caption{CELP self-supervised pretraining}
\label{alg:celp}

\textbf{Input}: EEG corpus $\mathcal{X}$; online/target encoders
$f_\theta, f_{\bar\theta}$; predictor $g_\phi$; momentum schedule
$\{m_k\}$\\

\textbf{Output}: Pretrained encoder $f_\theta$

\begin{algorithmic}[1]

\STATE $\bar\theta \leftarrow \theta$

\FOR{$k = 1$ \TO $K$}

    \STATE Sample batch $X \sim \mathcal{X}$; embed tokens
    (temporal $+$ spectral) $\rightarrow \tilde{Z}$

    \STATE Draw visible mask $\mathcal{M}_v$, prediction-mask set
    $\mathcal{M}=\{\mathcal{M}_p\}$
    \COMMENT{structured multi-block}

    \STATE $H_v \leftarrow
    f_\theta(\tilde{Z}_{\mathcal{M}_v})$
    \COMMENT{online: visible only}

    \STATE $H_t \leftarrow
    \mathrm{sg}\big(\mathrm{LN}(f_{\bar\theta}(\tilde{Z}))\big)$
    \COMMENT{target: full seq, stop-grad}

    \STATE $\hat{H}_{\mathcal{M}_p} \leftarrow
    g_\phi(H_v,\ \mathbf{m}+p_{\mathcal{M}_p})$
    \COMMENT{predict masked latents}

    \STATE $\mathcal{L} \leftarrow
    \frac{1}{|\mathcal{M}|}
    \sum_{\mathcal{M}_p \in \mathcal{M}}
    \frac{1}{|\mathcal{M}_p|}
    \sum_{i\in\mathcal{M}_p}
    \rho(\hat{h}_i, h_{t,i})$
    \COMMENT{Smooth $L_1$}

    \STATE Update $(\theta,\phi)$ via $\nabla\mathcal{L}$;\quad
    $\bar\theta \leftarrow
    m_k\bar\theta+(1-m_k)\theta$
    \COMMENT{EMA}

\ENDFOR

\STATE \textbf{return} $f_\theta$
\COMMENT{discard $g_\phi$}

\end{algorithmic}
\end{algorithm}

\subsection{CELP Encoder Pretraining}

Each iteration performs one gradient update of the online encoder
$f_\theta$ and predictor $g_\phi$, followed by an EMA update of the target
encoder $f_{\bar\theta}$. Three key architectural choices underpin this
strategy. First, the targets originate from the \emph{momentum} encoder
under a stop-gradient operation rather than from $f_\theta$ itself; this
asymmetry helps prevent trivial representation collapse. Second, the loss
is a Smooth-$L_1$ discrepancy computed over layer-normalized
\emph{latents}, ensuring the model avoids overfitting to high-frequency
waveform details. Third, the predictor is exclusively used to solve the
masked-prediction task and is discarded post-pretraining, yielding a
streamlined encoder for downstream applications.

\begin{algorithm}[!t]
\caption{Inference by semantic matching}
\label{alg:infer}

\textbf{Input}: EEG $X_{\mathrm{EEG}}$; task $k$ with instruction $s_k$
and candidate labels $\mathcal{C}_k$ (with verbalizations $\{\ell_c\}$);
frozen encoder $f_\theta$; semantic projector $P_h$; EEG token projector
$P_u$; queries $\{Q_m\}_{m=1}^{M}$; predictor $G$; head $P_a$;
text encoder $E_{\mathrm{text}}$\\

\textbf{Output}: Predicted label $\hat{y}$

\begin{algorithmic}[1]

\STATE $H_{\mathrm{EEG}} \leftarrow
\mathrm{reshape}\big(f_\theta(\tilde{Z}(X_{\mathrm{EEG}}))\big)$

\STATE $h \leftarrow \mathrm{Flatten}(H_{\mathrm{EEG}})$

\FOR{$m = 1$ \TO $M$}

    \STATE $\tilde{s}_m \leftarrow
    \mathrm{Pool}_{\mathrm{tok}}
    \big(\mathrm{CrossAttn}(Q_m, h, h)\big)$

    \STATE $s_m \leftarrow P_h(\tilde{s}_m)$
    \COMMENT{project semantic summary into LM space}

\ENDFOR

\STATE $S_q \leftarrow [s_1;\dots;s_M]$

\STATE $u \leftarrow P_u(h)$
\COMMENT{project EEG tokens into LM space}

\STATE $T_k \leftarrow \mathrm{Embed}_{\mathrm{LM}}(s_k)$

\STATE $H^{(k)} \leftarrow G([T_k; S_q; u])$

\STATE $H_q^{(k)} \leftarrow H^{(k)}_{\mathrm{query}}$
\COMMENT{states at semantic-query positions}

\STATE $\hat{z} \leftarrow
\mathrm{norm}\big(P_a(\mathrm{Pool}_{\mathrm{qry}}(H_q^{(k)}))\big)$
\COMMENT{answer embedding}

\FOR{\textbf{each} label $c \in \mathcal{C}_k$}

    \STATE $z_c \leftarrow
    \mathrm{norm}(E_{\mathrm{text}}(\ell_c))$
    \COMMENT{verbalized label}

\ENDFOR

\STATE $\hat{y} \leftarrow
\arg\max_{c\in\mathcal{C}_k}\ \hat{z}^{\top}z_c$

\STATE \textbf{return} $\hat{y}$

\end{algorithmic}
\end{algorithm}

\subsection{Inference by semantic matching}

Inference for each EEG sample is executed via a single forward pass through
the EEG-conditioned prediction model without autoregressive generation.
The frozen encoder and multi-query module generate a unified answer
embedding $\hat{z}$; each candidate label is encoded using the frozen text
encoder, and the final prediction corresponds to the nearest label under
cosine similarity. This formulation provides two key advantages: the
candidate set $\mathcal{C}_k$ serves as an \emph{input}, allowing the model
to accommodate different candidate label sets without adding task-specific
classification heads; furthermore, since scoring relies solely on
dot-product operations against label embeddings, the computational overhead
of decoding is negligible.

\section{Additional Results on Downstream Tasks}
\label{app:full_results}

\subsection{Abnormal Detection}
\subsubsection{Results on TUAB.}
Table~\ref{tab:tuab_results} presents the abnormal EEG detection performance on the TUAB dataset. BLPM achieves the best results across all three metrics, with a B-Acc of 82.32\%, an AUPRC of 89.71\%, and an AUROC of 90.36\%. Compared with CBraMod, the strongest baseline, BLPM improves B-Acc, AUPRC, and AUROC by 1.42, 1.06, and 1.52 percentage points, respectively. Abnormal EEG patterns can vary substantially not only across disease types but also according to the physiological and pathological characteristics of individual patients. Continuous semantic embedding prediction allows these heterogeneous abnormal patterns to be organized around the semantics of abnormality within a shared embedding space. This demonstrates that such semantic organization can contribute to a more stable separation of diverse abnormal EEG recordings from normal EEG activity.

\begin{table*}[t]
\centering
\begin{tabular}{lccc}
\toprule
\textbf{Model}
& \textbf{B-Acc}
& \textbf{AUPRC}
& \textbf{AUROC} \\
\midrule

\multicolumn{4}{@{}l}{
\textit{Task-specific models} (trained from scratch)
} \\

EEGNet~\citep{lawhern2018eegnet}
& 79.86$\pm$0.36
& 87.71$\pm$0.44
& 87.81$\pm$0.33 \\

EEGConformer~\citep{song2023eegconformer}
& 78.97$\pm$0.43
& 87.55$\pm$0.50
& 87.38$\pm$0.51 \\

\midrule

\multicolumn{4}{@{}l}{
\textit{EEG foundation models}
(finetuned with a task-specific head)
} \\

BIOT~\citep{yang2023biot}
& 75.69$\pm$0.24
& 80.07$\pm$1.53
& 82.19$\pm$0.69 \\

LaBraM~\citep{jiang2024labram}
& 77.23$\pm$1.49
& 82.48$\pm$1.23
& 84.22$\pm$1.35 \\

CBraMod~\citep{wang2025cbramod}
& \underline{80.90$\pm$0.26}
& \underline{88.65$\pm$0.36}
& \underline{88.84$\pm$0.30} \\

REVE~\citep{elouahidi2025reve}
& 78.13$\pm$1.51
& 80.57$\pm$2.00
& 83.10$\pm$1.72 \\

LUNA~\citep{doner2025luna}
& 80.19$\pm$0.60
& 88.37$\pm$0.49
& 88.36$\pm$0.33 \\

\midrule

\multicolumn{4}{@{}l}{
\textit{Unified model}
(a single model for all tasks, no task-specific head)
} \\

\textbf{BLPM (Ours)}
& \textbf{82.32$\pm$0.23}
& \textbf{89.71$\pm$0.31}
& \textbf{90.36$\pm$0.58} \\

\bottomrule
\end{tabular}

\caption{
Abnormal detection results on TUAB, reported as mean $\pm$ standard deviation over five seeds. All metrics are reported as percentages (\%). The best and second-best results are marked in bold and underlined, respectively.
}
\label{tab:tuab_results}
\end{table*}

\subsection{Motor Imagery Classification}
\subsubsection{Results on PhysioNet-MI.}

\begin{table*}[t]
\centering
\begin{tabular}{lccc}
\toprule
\textbf{Model}
& \textbf{B-Acc}
& \textbf{Cohen's $\kappa$}
& \textbf{F1-W} \\
\midrule

\multicolumn{4}{@{}l}{
\textit{Task-specific models} (trained from scratch)
} \\

EEGNet~\citep{lawhern2018eegnet}
& 58.25$\pm$0.46
& 44.34$\pm$0.61
& 57.65$\pm$0.56 \\

EEGConformer~\citep{song2023eegconformer}
& 57.02$\pm$1.08
& 43.05$\pm$1.71
& 57.05$\pm$1.10 \\

\midrule

\multicolumn{4}{@{}l}{
\textit{EEG foundation models}
(finetuned with a task-specific head)
} \\

BIOT~\citep{yang2023biot}
& 30.77$\pm$2.23
& 7.69$\pm$2.98
& 29.81$\pm$2.17 \\

LaBraM~\citep{jiang2024labram}
& 59.33$\pm$0.73
& 47.74$\pm$1.51
& 61.06$\pm$1.25 \\

CBraMod~\citep{wang2025cbramod}
& 56.61$\pm$1.21
& 40.95$\pm$1.08
& 55.35$\pm$1.12 \\

REVE~\citep{elouahidi2025reve}
& \textbf{69.40$\pm$0.92}
& \textbf{58.93$\pm$1.87}
& \textbf{69.32$\pm$1.21} \\

LUNA~\citep{doner2025luna}
& 61.72$\pm$0.62
& 49.37$\pm$1.49
& 61.95$\pm$1.21 \\

\midrule

\multicolumn{4}{@{}l}{
\textit{Unified model}
(a single model for all tasks, no task-specific head)
} \\

\textbf{BLPM (Ours)}
& \underline{68.22$\pm$0.71}
& \underline{57.85$\pm$1.23}
& \underline{68.35$\pm$1.12} \\

\bottomrule
\end{tabular}

\caption{
Motor imagery classification results on PhysioNet-MI, reported as mean $\pm$ standard deviation over five seeds. All metrics are reported as percentages (\%). The best and second-best results are marked in bold and underlined, respectively.
}
\label{tab:physionet_mi_results}
\end{table*}

Table~\ref{tab:physionet_mi_results} presents the motor imagery classification results on PhysioNet-MI. BLPM achieves a B-Acc of 68.22\%, Cohen's $\kappa$ of 57.85\%, and F1-W of 68.35\%, ranking second across all three metrics. BLPM follows REVE by 1.18, 1.08, and 0.97 percentage points in B-Acc, Cohen's $\kappa$, and F1-W, respectively, demonstrating consistently competitive performance across all three metrics. In particular, BLPM substantially outperforms the task-specific models, improving B-Acc by 9.97 percentage points over EEGNet and by 11.20 percentage points over EEGConformer. Motor imagery classification requires distinguishing subtle class-dependent changes in sensorimotor activity, including spatially distributed cortical responses and variations in oscillatory dynamics associated with different imagined movements. These patterns can also vary considerably across subjects, making it important to preserve fine-grained neural differences while learning representations that remain consistent across individuals. The strong performance of BLPM suggests that continuous semantic embedding prediction can retain these discriminative sensorimotor characteristics and organize heterogeneous neural responses corresponding to the same imagined movement around a shared class-level representation. This enables BLPM to effectively distinguish different motor imagery classes despite substantial inter-subject variability while maintaining performance comparable to the strongest baseline.

\subsection{Emotion Recognition}
\subsubsection{Results on FACED.}
\begin{table*}[t]
\centering
\begin{tabular}{lccc}
\toprule
\textbf{Model}
& \textbf{B-Acc}
& \textbf{Cohen's $\kappa$}
& \textbf{F1-W} \\
\midrule

\multicolumn{4}{@{}l}{
\textit{Task-specific models} (trained from scratch)
} \\

EEGNet~\citep{lawhern2018eegnet}
& 23.36$\pm$0.69
& 13.48$\pm$0.78
& 22.10$\pm$0.74 \\

EEGConformer~\citep{song2023eegconformer}
& 29.01$\pm$0.52
& 19.65$\pm$0.59
& 27.51$\pm$0.62 \\

\midrule

\multicolumn{4}{@{}l}{
\textit{EEG foundation models}
(finetuned with a task-specific head)
} \\

BIOT~\citep{yang2023biot}
& 11.21$\pm$0.29
& 0.22$\pm$0.72
& 8.29$\pm$1.91 \\

LaBraM~\citep{jiang2024labram}
& 26.84$\pm$0.94
& 18.01$\pm$1.30
& 26.56$\pm$1.53 \\

CBraMod~\citep{wang2025cbramod}
& 29.41$\pm$1.52
& 21.36$\pm$1.13
& 29.23$\pm$0.89 \\

REVE~\citep{elouahidi2025reve}
& \underline{32.03$\pm$1.23}
& \underline{23.90$\pm$0.99}
& \underline{31.83$\pm$1.18} \\

LUNA~\citep{doner2025luna}
& 29.30$\pm$11.76
& 22.37$\pm$1.50
& 30.22$\pm$1.42 \\

\midrule

\multicolumn{4}{@{}l}{
\textit{Unified model}
(a single model for all tasks, no task-specific head)
} \\

\textbf{BLPM (Ours)}
& \textbf{34.32$\pm$0.61}
& \textbf{24.21$\pm$0.31}
& \textbf{33.07$\pm$1.22} \\

\bottomrule
\end{tabular}

\caption{
Emotion recognition results on FACED, reported as mean $\pm$ standard deviation over five seeds. All metrics are reported as percentages (\%). The best and second-best results are marked in bold and underlined, respectively.
}
\label{tab:faced_results}
\end{table*}

Table~\ref{tab:faced_results} indicates the emotion recognition performance on the FACED dataset. BLPM achieves the best results across all three metrics, with a B-Acc of 34.32\%, a Cohen's $\kappa$ of 24.21\%, and an F1-W of 33.07\%. Compared with REVE, the strongest baseline, BLPM improves B-Acc, Cohen's $\kappa$, and F1-W by 2.29, 0.31, and 1.24 percentage points, respectively. Emotion recognition on FACED is particularly challenging because it requires distinguishing among nine emotion categories whose EEG responses may exhibit subtle and overlapping affective patterns. Despite this fine-grained nine-class setting, BLPM consistently outperforms the competing models. Rather than relying solely on conventional task-specific classification boundaries, BLPM structures neural patterns associated with different affective states according to the semantic relationships among emotion categories, thereby learning more discriminative and transferable representations for fine-grained emotion recognition.

\subsection{Sleep Staging}
\subsubsection{Results on HMC.}

\begin{table*}[t]
\centering
\begin{tabular}{lccc}
\toprule
\textbf{Model}
& \textbf{B-Acc}
& \textbf{Cohen's $\kappa$}
& \textbf{F1-W} \\
\midrule

\multicolumn{4}{@{}l}{
\textit{Task-specific models} (trained from scratch)
} \\

EEGNet~\citep{lawhern2018eegnet}
& 70.89$\pm$0.81
& 59.22$\pm$2.01
& 66.89$\pm$2.08 \\

EEGConformer~\citep{song2023eegconformer}
& 72.29$\pm$1.61
& 61.67$\pm$1.48
& 68.59$\pm$1.62 \\

\midrule

\multicolumn{4}{@{}l}{
\textit{EEG foundation models}
(finetuned with a task-specific head)
} \\

BIOT~\citep{yang2023biot}
& 71.54$\pm$1.12
& 62.41$\pm$2.16
& 70.51$\pm$1.85 \\

LaBraM~\citep{jiang2024labram}
& 75.34$\pm$0.89
& \underline{67.71$\pm$1.21}
& \underline{75.19$\pm$1.07} \\

CBraMod~\citep{wang2025cbramod}
& 76.01$\pm$0.92
& 66.56$\pm$0.89
& 73.34$\pm$0.73 \\

REVE~\citep{elouahidi2025reve}
& \underline{76.54$\pm$0.84}
& 67.10$\pm$1.10
& 74.79$\pm$1.24 \\

LUNA~\citep{doner2025luna}
& 74.82$\pm$0.81
& 66.39$\pm$1.14
& 73.81$\pm$0.91 \\

\midrule

\multicolumn{4}{@{}l}{
\textit{Unified model}
(a single model for all tasks, no task-specific head)
} \\

\textbf{BLPM (Ours)}
& \textbf{77.12$\pm$0.64}
& \textbf{69.07$\pm$1.85}
& \textbf{76.98$\pm$0.67}\\

\bottomrule
\end{tabular}

\caption{
Sleep staging results on HMC, reported as mean $\pm$ standard deviation over five seeds. All metrics are reported as percentages (\%). The best and second-best results are marked in bold and underlined, respectively.
}
\label{tab:hmc_multimetric}
\end{table*}

Table~\ref{tab:hmc_multimetric} presents the sleep staging performance on the HMC dataset. BLPM achieves the best results across all three metrics, with a B-Acc of 77.12\%, Cohen's $\kappa$ of 69.07\%, and F1-W of 76.98\%. Compared with the strongest baseline for each metric, BLPM improves B-Acc by 0.58 percentage points over REVE, while improving Cohen's $\kappa$ and F1-W by 1.36 and 1.79 percentage points over LaBraM, respectively. Sleep staging requires distinguishing between sleep stages that exhibit gradual transitions and partially overlapping electrophysiological characteristics. Nevertheless, BLPM consistently achieves strong and balanced performance across stages. These results indicate that BLPM effectively organizes stage-related neural patterns in its representation space, enabling it to distinguish subtle differences between sleep states without relying on a task-specific classification head.

\section{Visualization Results}
To qualitatively examine the effect of semantic alignment, we visualize the learned representations using t-SNE. We analyze both the global distribution across all downstream datasets and the class-level structure within representative tasks to assess how semantic alignment reshapes the representation space at both the dataset and class levels.

\subsection{Global Representation Visualization}

\begin{figure*}[t]
\centering
\includegraphics[width=\textwidth]{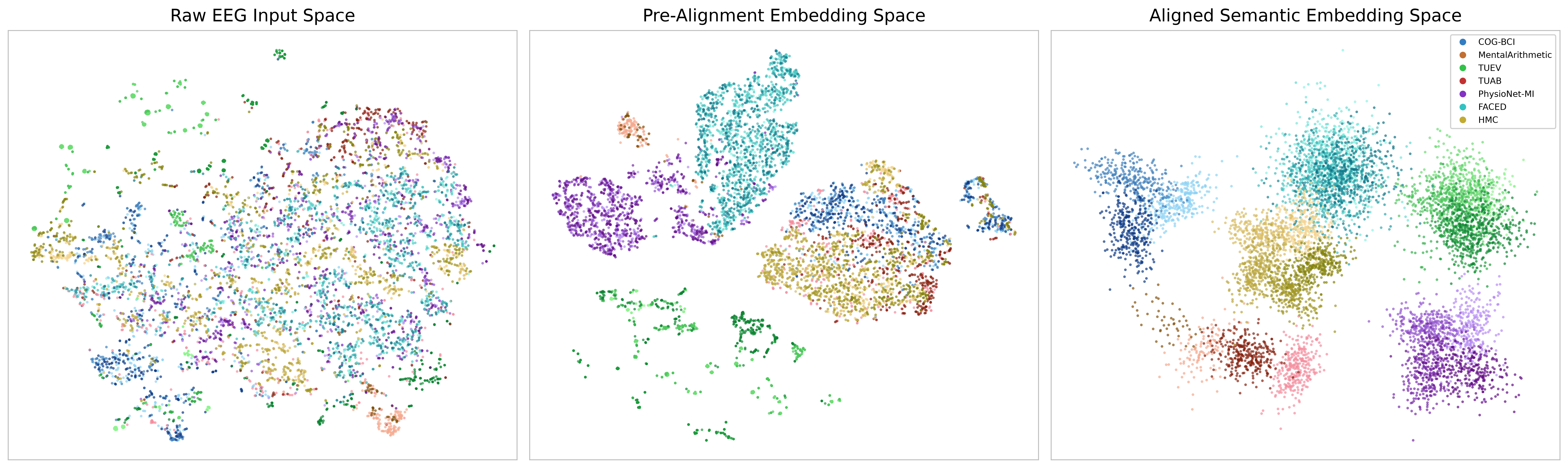}
\caption{Global t-SNE visualization across the seven downstream datasets.}
\label{fig:global_alignment}
\end{figure*}

Figure~\ref{fig:global_alignment} compares the distributions of samples from the seven downstream datasets in the input space, the pretrained EEG representation space, and the aligned semantic embedding space. In the input space, samples from different datasets are broadly intermixed, reflecting substantial heterogeneity in recording configurations and signal characteristics. The pretrained CELP encoder produces a more structured distribution, although considerable overlap remains among datasets. After semantic alignment, samples from the same dataset form more coherent regions, while the distributions of different datasets become more systematically organized in the shared embedding space. These observations illustrate how semantic alignment changes the organization of the learned representation space while preserving dataset-specific characteristics.

\subsection{Task-Specific Representation Visualization}

\begin{figure*}[t]
\centering
\includegraphics[
    width=\textwidth,
    height=0.78\textheight,
    keepaspectratio
]{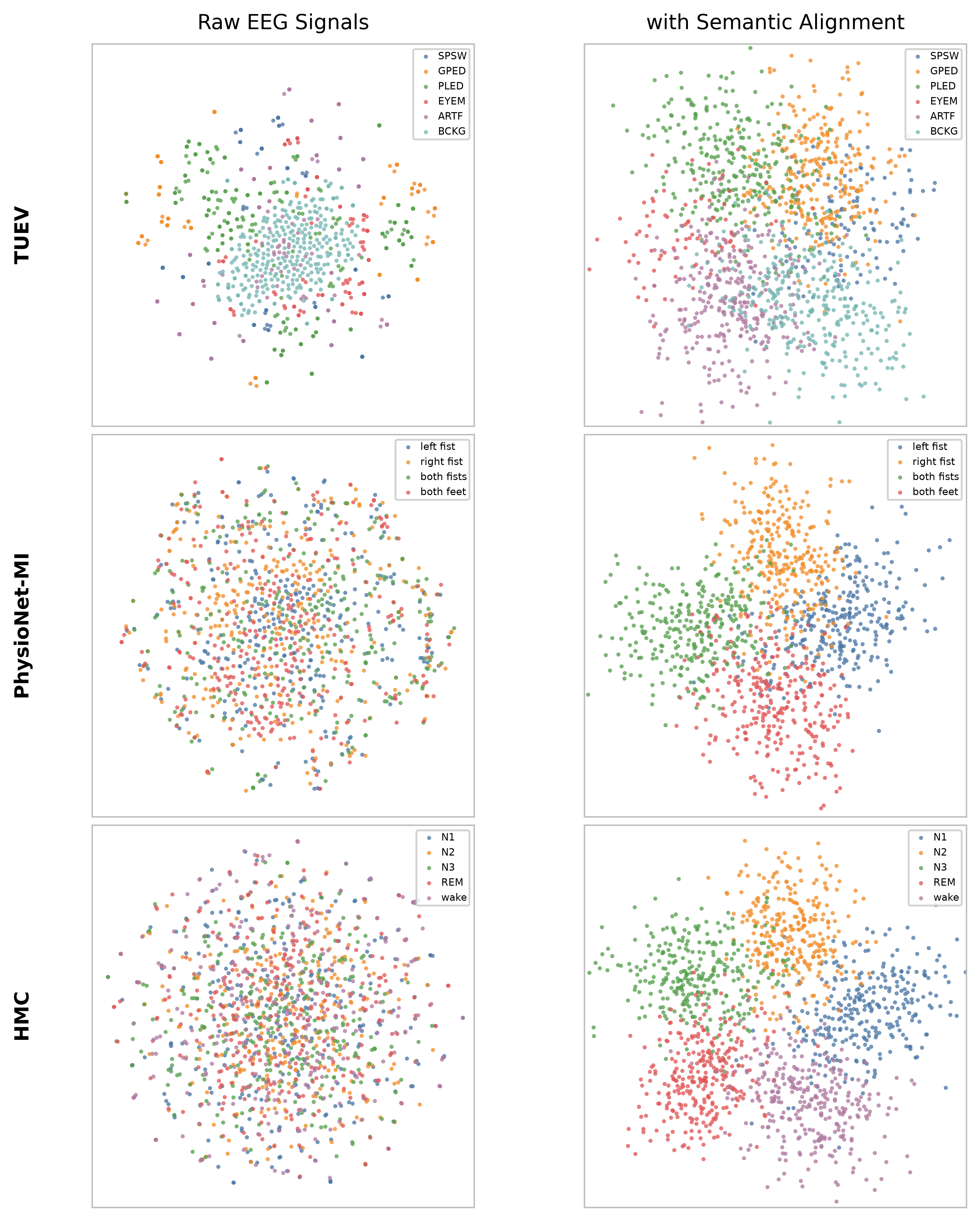}
\caption{Task-specific t-SNE visualizations for TUEV, PhysioNet-MI, and HMC. The left and right columns show the representation distributions before and after semantic alignment, respectively.}
\label{fig:task_alignment}
\end{figure*}

Figure~\ref{fig:task_alignment} visualizes the representations before and after semantic alignment for three representative tasks: clinical event classification on TUEV, motor imagery classification on PhysioNet-MI, and sleep-stage classification on HMC. Before alignment, samples from different classes are substantially intermixed across all three tasks. After alignment, the representations exhibit clearer class-dependent organization, with samples from the same class forming more coherent regions. Although partial overlap remains between physiologically or morphologically related classes, the overall distributions indicate that semantic alignment enhances class-discriminative structure across heterogeneous EEG decoding tasks.

\end{document}